\documentclass{article}

\usepackage{amsmath,amsfonts,bm}

\def\eqref#1{equation~\ref{#1}}
\def\1{\bm{1}}

\def\vs{{\bm{s}}}

\def\mA{{\bm{A}}}

\def\mS{{\bm{S}}}

\def\mV{{\bm{V}}}
\def\mW{{\bm{W}}}
\def\mX{{\bm{X}}}

\DeclareMathAlphabet{\mathsfit}{\encodingdefault}{\sfdefault}{m}{sl}
\SetMathAlphabet{\mathsfit}{bold}{\encodingdefault}{\sfdefault}{bx}{n}

\usepackage{microtype}
\usepackage{graphicx}
\usepackage{subfigure}
\usepackage{booktabs} % for professional tables

\usepackage{hyperref}

\usepackage[accepted]{icml2023}

\usepackage{amsmath}
\usepackage{amssymb}
\usepackage{mathtools}
\usepackage{amsthm}
\usepackage{makecell}

\usepackage[capitalize,noabbrev]{cleveref}

\theoremstyle{plain}
\newtheorem{theorem}{Theorem}[section]
\newtheorem{proposition}[theorem]{Proposition}

\theoremstyle{definition}

\theoremstyle{remark}

\usepackage[textsize=tiny]{todonotes}

\icmltitlerunning{FlexRound: Learnable Rounding based on Element-wise Division for Post-Training Quantization}

\begin{document}

\twocolumn[
\icmltitle{FlexRound: Learnable Rounding based on Element-wise Division \\ for Post-Training Quantization}



\icmlsetsymbol{equal}{*}

\begin{icmlauthorlist}
\icmlauthor{Jung Hyun Lee}{equal,comp}
\icmlauthor{Jeonghoon Kim}{equal,comp}
\icmlauthor{Se Jung Kwon}{comp}
\icmlauthor{Dongsoo Lee}{comp}
\end{icmlauthorlist}

\icmlaffiliation{comp}{NAVER Cloud, Seongnam, South Korea}

\icmlcorrespondingauthor{Jung Hyun Lee}{onliwad101@gmail.com}
\icmlcorrespondingauthor{Jeonghoon Kim}{jeonghoon.samuel@gmail.com}

\icmlkeywords{Machine Learning, ICML}

\vskip 0.3in
]



\printAffiliationsAndNotice{\icmlEqualContribution} % otherwise use the standard text.

\begin{abstract}
Post-training quantization (PTQ) has been gaining popularity for the deployment of deep neural networks on resource-limited devices since unlike quantization-aware training, neither a full training dataset nor end-to-end training is required at all. As PTQ schemes based on reconstructing each layer or block output turn out to be effective to enhance quantized model performance, recent works have developed algorithms to devise and learn a new weight-rounding scheme so as to better reconstruct each layer or block output.
% has become popular in PTQ, it is crucial how to devise and learn a new rounding mechanism.
%We notice that such new rounding schemes are established on element-wise addition. %, which can limit the exploration of rounding search space.
In this work, we propose a simple yet effective new weight-rounding mechanism for PTQ, coined \emph{FlexRound}, based on element-wise division instead of typical element-wise addition such that FlexRound enables jointly learning a common quantization grid size as well as a different scale for each pre-trained weight. Thanks to the reciprocal rule of derivatives induced by element-wise division, FlexRound is inherently able to exploit pre-trained weights when updating their corresponding scales, and thus, flexibly quantize pre-trained weights depending on their magnitudes. We empirically validate the efficacy of FlexRound on a wide range of models and tasks. To the best of our knowledge, our work is the first to carry out comprehensive experiments on not only image classification and natural language understanding but also natural language generation. Moreover, we demonstrate, for the first time, that large language models can be efficiently quantized, with only a negligible impact on performance compared to half-precision baselines, achieved by reconstructing the output in a block-by-block manner.
%  Our code is available at \url{https://github.com/clovaai/FlexRound}.
% we show that FlexRound can quantize large language models under a per-channel uniform PTQ reconstruction with only marginal performance degradation compared to the half-precision baselines. 
Our code is available at \url{https://github.com/onliwad101/FlexRound_LRQ}.
\end{abstract}

\section{Introduction}

In recent years, deep neural networks have achieved unprecedented success across a wide variety of domains such as computer vision, natural language processing, and automatic speech recognition.
Unfortunately, as these networks continue to improve and surpass human-level performance, the computational resources and memory usage required also increases as the architecture becomes more complex.
%Recent years have witnessed the unprecedented success of deep neural networks in a wide variety of domains including computer vision, natural language processing, automatic speech recognition, and so on.
% tasks including image classification, object detection, machine translation, natural language understanding, natural language generation, speech recognition, and so on. 
%Even if state-of-the-art deep neural networks surpass human-level performance, these neural networks cannot help requiring more and more computation cost and memory usage as networks become deeper and wider.
To reduce the model size and accelerate inference operations, many researchers have attempted diverse compression techniques such as network quantization \citep{courbariaux2016binarized} and network pruning \citep{han2016deep}. In this paper, we concentrate on network quantization due to the advantage that INT4 or INT8 quantization allows us to accelerate quantized neural networks using off-the-shelf accelerators such as the NVIDIA A100 Tensor Core GPU \citep{wu2020integer} or ARM Cortex MCUs \citep{kim2021performance}.

Network quantization techniques can be broadly divided into two categories: quantization-aware training (QAT) and post-training quantization (PTQ). QAT is a method where the quantization of the networks is incorporated during the trianing process, as proposed by various research works such as \citet{jung2019learning, jain2019tqt, zhao2020linear, esser2020learned, lee2021cluster}. We note that QAT results in a marginal performance difference between the full-precision and quantized versions of the neural network. Yet, QAT requires end-to-end retraining or fine-tuning on a full training dataset, which often causes an enormous amount of time and resources to obtain a quantized neural network with competitive performance. Furthermore, a whole training dataset may not be available due to data privacy issues or demands to utilize legacy models. Such drawbacks of QAT are the reasons why researchers recently pay more attention to PTQ \citep{zhao2019improving, wang2020towards, nahshan2021loss} that needs neither a full training dataset nor end-to-end learning at all. 
% only a small unlabelled dataset and does not require end-to-end learning at all. 

PTQ had been initially performed via rounding-to-nearest by minimizing the quantization error in the parameter space. However, this approach suffers from severe performance degradation. 
Since it is reported that the loss degradation resulting from quantization can be approximated as the second-order error in Taylor Expansion by viewing quantized weights as perturbed weights, \citet{nagel2020adaround} and \citet{li2021brecq} substantiate that reconstructing each output of layer or block is equivalent to minimizing the approximation of loss degradation resulting from quantization under some assumptions. Accordingly, recent works \citep{nagel2020adaround, li2021brecq, hubara2021adaquant, wei2022qdrop} have suggested to reconstruct each output of layer or block by devising and learning a new weight-rounding scheme, deviating from rounding-to-nearest, as an effort to preserve the performance of a full-precision model even after PTQ.
However, all those new rounding schemes designed in existing studies either round or quantize pre-trained weights adaptively via element-wise addition.
% Recent works \citep{nagel2020adaround, hubara2021adaquant, li2021brecq, wei2022qdrop} have made an effort to reconstruct each output of layer or block by learning a new rounding scheme, deviating from rounding-to-nearest. Considering that the loss degradation resulting from quantization can be approximated as the second-order error in Taylor Expansion, they have concentrated on the minimization of the approximation of loss degradation resulting from quantization in order to make sure to preserve the performance of a full-precision model.  However, all new rounding schemes in existing studies either round or quantize pre-trained weights adaptively via element-wise addition.
% are based on element-wise addition in deciding to which quantization grid to discretize each pre-trained weight.
% (to weights or quantized weights), which make it hard to fully leverage the information of pre-trained weights when deciding where to quantize each pre-trained weight.

We propose a novel post-training weight quantization method, called FlexRound, which departs from the typical element-wise addition approaches and instead employs an element-wise division perspective.
%Our method is both simple and effective in its ability to flexibly quantize pre-trained weights.
By jointly learning a common quantization grid size and the division factor for pre-trained weights, FlexRound offers a new approach to PTQ.
%Changing the perspective of a new rounding policy from element-wise addition to element-wise division, we propose a simple yet effective post-training weight quantization method called FlexRound, which \textcolor{blue}{can} flexibly quantize pre-trained weights by \textcolor{blue}{jointly} learning \textcolor{blue}{not just} how much each pre-trained weight should be divided by \textcolor{blue}{but also a common quantization grid size}.
% updating an individual scale for every pre-trained weight as well as a common quantization grid size. 
% In other words, through learning how much each pre-trained weight is divided by, FlexRound can flexibly quantize pre-trained weights.
% which learns each denominator of the fraction inside a rounding function for each pre-trained weight as well as the scaling factor outside a rounding function. \textcolor{blue}{In other words, through dividing each pre-trained weight by a different scale inside a rounding function, FlexRound can flexibly quantize pre-trained weights while sharing a common scaling factor outside a round function.}
% In other words, FlexRound flexibly quantizes pre-trained weights by dividing each pre-trained weight by a different value inside a rounding function while sharing a common scaling factor outside a round function. 
Interestingly, thanks to the reciprocal rule of derivatives induced by element-wise division, FlexRound can inherently leverage pre-trained weights when updating an individual scale for each pre-trained weight. Specifically, we corroborate that a relatively wider range of discrete values needs to be explored when quantizing pre-trained weights of large magnitude. The rationale behind such an approach is that the magnitude of a weight can be interpreted as its relative importance within the network. Given that weights of larger magnitude have a greater impact on the network's performance than those of smaller magnitude, as demonstrated by research such as \citep{han2016deep}, to maintain the performance of a pre-trained model even after quantization,
% it is crucial to retain the knowledge of important weights even after quantization so as to maintain the performance of a pre-trained model, 
it is important to relax the constraints associated with quantizing weights of large absolute value compared to those of small absolute value (i.e., important weights can be quantized to one of not only its two nearest discrete values but also to discrete values further away from it). Accordingly, FlexRound can quantize pre-trained weights flexibly depending on their own magnitudes, % importance, 
thereby leading to better performance.

Our contributions are threefold:
\begin{itemize}
    \item We propose FlexRound as a new rounding scheme for post-training weight quantization based on the principle of element-wise division in order to allow for jointly learning not only a separate scale for every pre-trained weight but also a common quantization grid size across a group (e.g., a channel or a layer).
% 	We propose a new post-training weight quantization named FlexRound that flexibly discretizes pre-trained weights by learning how much each pre-trained weight is divided by inside a rounding function.
% 	based on element-wise division instead of element-wise addition. 
% 	We for the first time devise a new rounding scheme for PTQ in the view of element-wise division.
    \item We theoretically and empirically demonstrate that such a new rounding scheme based on element-wise division takes into consideration the magnitude of pre-trained weights when updating their corresponding scales so that FlexRound can quantize pre-trained weights of large magnitude (i.e., important pre-trained weights) more flexibly than rounding either up or down only.
% 	in updating their corresponding denominators inside a rounding function.
% 	take an advantage of the knowledge of a pre-trained model when quantizing pre-trained weights.
% 	We demonstrate that such a new rounding scheme via element-wise division enables FlexRound to decide where to quantize pre-trained weights by taking an advantage of the knowledge of a pre-trained model.
    \item To the best of our knowledge, we are the first to perform extensive experiments in a per-tensor uniform PTQ setting on 
% 	various tasks including 
    natural language generation as well as image classification and natural language understanding, using numerous models such as ResNet, MobileNetV2, BERT, GPT-Neo, OPT, and GPT-2. We also, for the first time, conduct the uniform PTQ reconstruction for large language models like LLaMA on both common sense reasoning and causal language modeling tasks. % in a per-channel uniform PTQ setting % the effectiveness of FlexRound
\end{itemize}

\begin{figure*}
    % \vskip -12pt
    \vskip -0.05in
    \centering
    \subfigure[A new rounding scheme based on element-wise division in a per-tensor uniform \newline PTQ setting. $s_1$ and $\mS$ are updated toward minimizing the reconstruction error, $\mathcal{L}$.] % Example of FlexRound scheme
    {
        \includegraphics[width=0.64\linewidth]{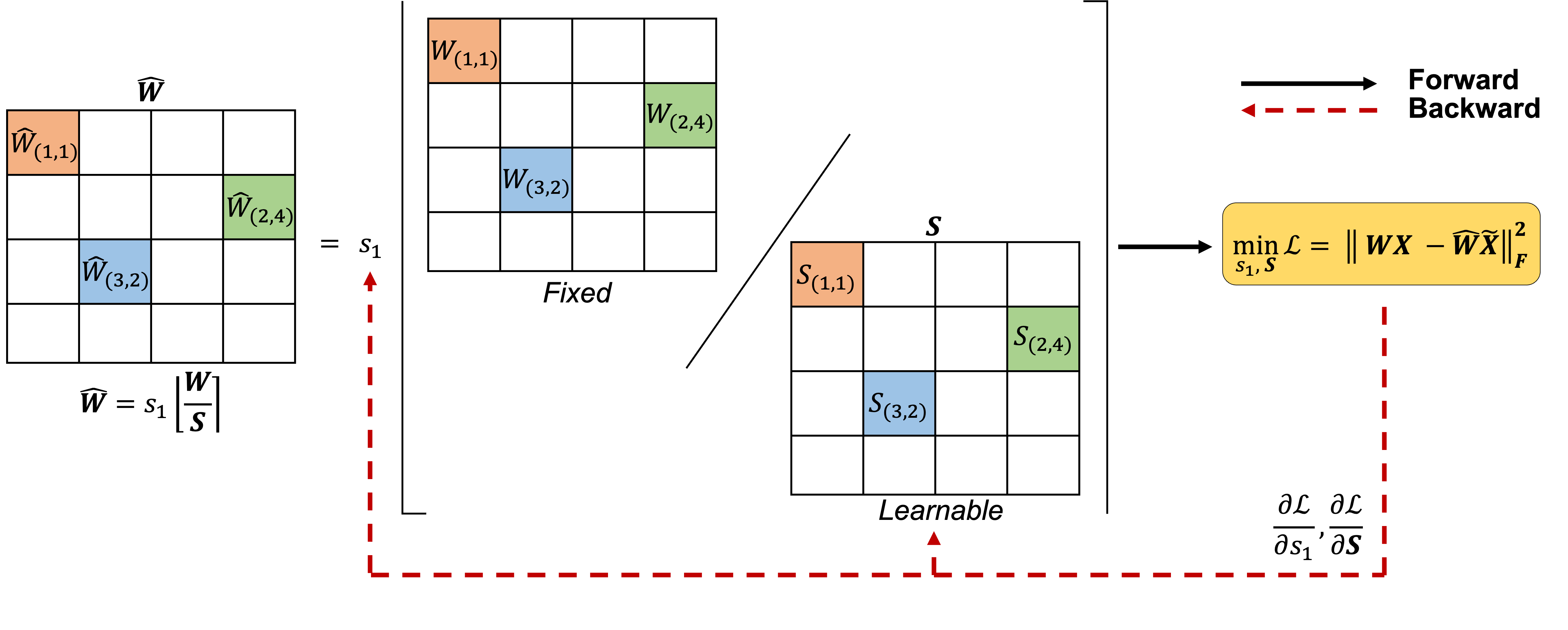}
        \label{fig1:a}
    }
    \subfigure[Rounding functions with learned parameters $s_1$ and $\mS$ as shown in (a).]
    {
        \includegraphics[width=0.32\linewidth]{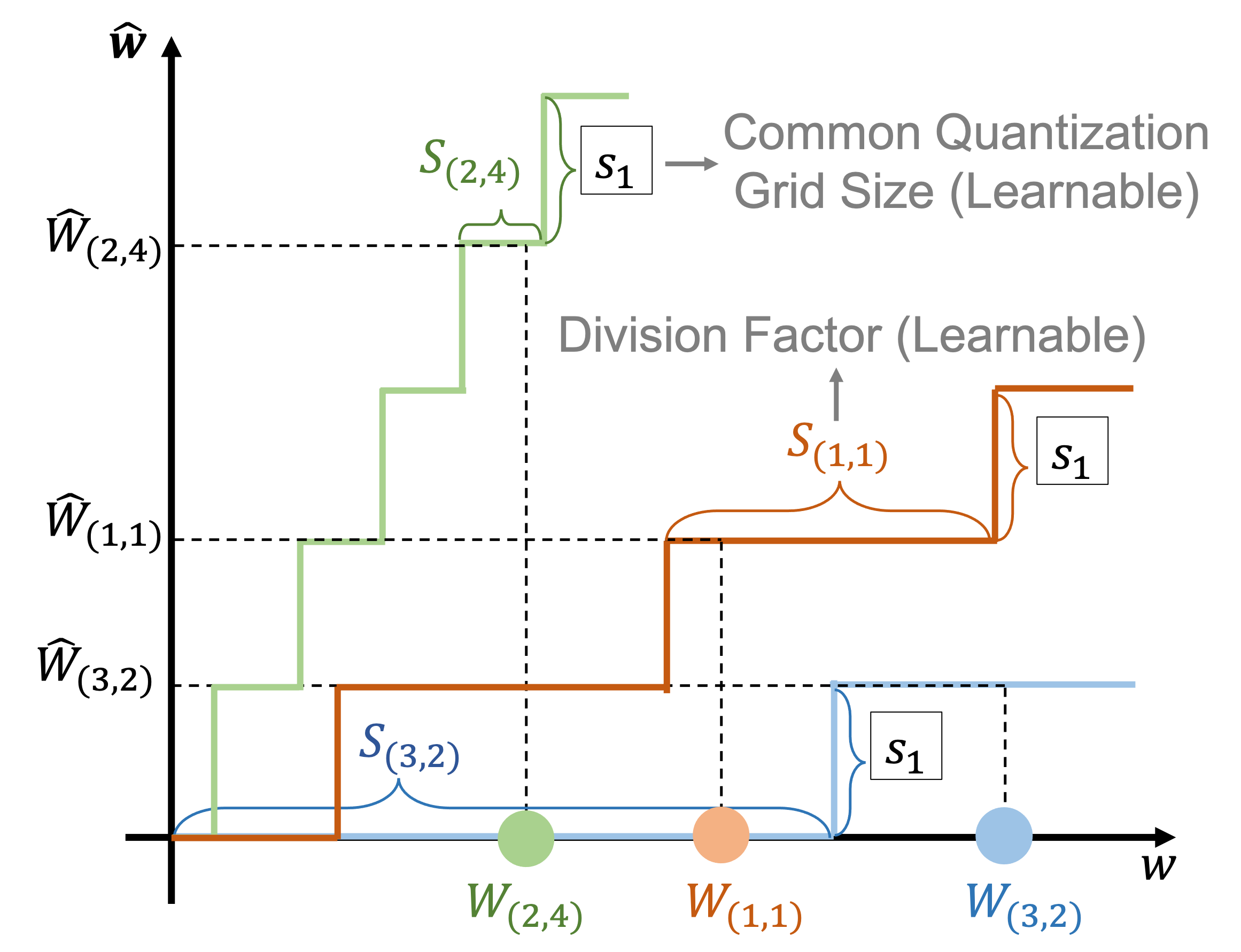}
        \label{fig1:b}
    }
    % \vskip -5pt
    \vskip -0.1in
    \caption{Illustration of FlexRound in the per-tensor uniform PTQ reconstruction. $s_1$ is a common quantization grid size across a layer, and $S_{(i, j)}$ is the division factor for a pre-trained weight $W_{(i, j)}$, both of which are positive and learnable. As shown in (b), with different learned $S_{(i, j)}$ via (a), FlexRound flexibly quantizes pre-trained weights by observing $W_{(2, 4)} < W_{(3, 2)}$ but $\widehat{W}_{(2, 4)} > \widehat{W}_{(3, 2)}$.}
    \label{fig:FlexRound}
    % \vskip -10pt
    \vskip -0.15in
\end{figure*}

\section{Related Work}\label{sec:related}
Recently, many researchers have attempted to quantize a wide range of models for various tasks such as computer vision and natural language understanding/generation without any (re)training. Outlier channel splitting (OCS) \citep{zhao2019improving} replicates channels entailing outliers, and then, halves outliers of those channels. Despite the fact that OCS explicitly addresses outliers, it still experiences severe accuracy degradation when both weights and activations are quantized to low-bit. As an alternative solution, \citet{wang2020towards} proposed Bit-Split that splits an integer into several bits and optimizes them separately.
While the performance of Bit-Split is comparable to that of a full-precision model in a low-bit setting, it may not be as effective for certain architectures such as MobileNetV2.
% While \citet{wang2020towards} demonstrated that the performance of the Bit-Split method is comparable to that of a full-precision model in a low-bit setting, it may not be as effective for certain architectures such as MobileNetV2.
%Although \citet{wang2020towards} showed that the performance of Bit-Split is close to that of a full-precision model in the low-bit setting, Bit-Split may not be effective for certain architectures, for example, MobileNetV2.

To overcome the limitations discussed above, \citet{nagel2020adaround} and \citet{hubara2021adaquant} minimize the mean squared error  (in a layer-by-layer fashion) between the full-precision layer's output and its quantized layer's output by inventing and learning a new weight-rounding mechanism dubbed as AdaRound and AdaQuant, respectively. As such a layer-wise reconstruction error minimization opens the door to $4$-bit PTQ regime, \citet{li2021brecq} proposed block-wise reconstruction, titled BRECQ, to consider cross-layer dependency along with the possibility of fully quantizing MobileNetV2 to $4$-bit. In addition to block-wise reconstruction, \citet{wei2022qdrop} proposed QDrop that drops the quantization of activations at random during the PTQ reconstruction to induce activation quantization to be synchronized with weight quantization. Both BRECQ and QDrop, however, are based on AdaRound that 
rounds weights only either up or down at most with a `fixed' quantization grid size.
% requires numerous hyper-parameters for regularization 
% while allowing for rounding either up or down only at most.
% AdaQuant does not need any hyper-parameters and
AdaQuant can simultaneously learn a quantization grid size and quantize weights adaptively, but incurs severe performance degradation when quantizing MobileNetV2 in low-bit regimes. % does not consider the magnitude of weights for quantization, which turns out to be important as we discuss later.
%, but it can excessively quantize weights of little importance or marginally quantize weights of large importance, thereby giving rise to accuracy degradation.

As another line of PTQ research, some PTQ techniques are exclusively specialized in quantizing language models such as BERT and GPT-like models. \citet{bondarenko2021understanding} first applied PTQ to BERT by introducing a per-embedding-group activation quantization scheme to deal with highly dynamic activation ranges. 
\citet{bai2021towards} studied the PTQ reconstruction in parallel for BERT.
\citet{yao2022zeroquant} proposed ZeroQuant that quantizes BERT and GPT-3 in a group-wise weight quantization manner driven by token-wise activation quantization via layer-by-layer knowledge distillation (while a dedicated CUDA kernel is required for ZeroQuant).  
\citet{dettmers2022llm} quantizes large language models (LLMs) like OPT with vector-wise weight quantization and mixed-precision decomposition with FP16 activations. To avoid the use of FP16 activations, \citet{xiao2022smoothquant} proposed SmoothQuant that shifts the difficulty of activation quantization to weight quantization, 
% the focus of quantization to the weights, 
allowing for INT8 quantization of both weights and activations in LLMs.
Unfortunately, both \citet{dettmers2022llm} and \citet{xiao2022smoothquant} assume that the outliers in activations would appear in a certain pattern.
% , which seems to be a disputable assumption for their deployment in practice.
% \textcolor{blue}{All those methods do not consider per-tensor weight quantization which can enable integer matrix-to-matrix multiplication API/function calls \citep{migacz20178}.} 
% \citet{migacz20178} (e.g., integer forms of cuBLAS and cuDNN that utilize tensor cores) for commercial GPUs.}

Most of the aforementioned PTQ studies are targeted to either vision models or language models only, not to both. 
% In addition, most of the experimental results in the above PTQ works are conducted via channel-wise/group-wise/vector-wise weight quantization.
% at the expense of reduced parallelism.
To the best of our knowledge, our work is the first to carry out extensive experiments on diverse tasks ranging from image classification and natural language understanding to natural language generation under a per-tensor uniform PTQ setting.
Additionally, we for the first time show that LLMs can be efficiently quantized, with only a minor impact on accuracy compared to half-precision baselines, attained by reconstructing each block output, without the assumption that the activation outliers would appear in a certain pattern.
%  can be quantized with only negligible accuracy degradation compared to the half-precision baselines by reconstructing each block output, without the assumption that the activation outliers would appear in a certain pattern.

\section{Methodology}\label{sec:methodolgy}

This section begins by introducing the notations used throughout the paper and the background of post-training quantization (PTQ). We then provide the concept and design of FlexRound for the uniform PTQ reconstruction method. We finally delve into the advantages of utilizing the principle of element-wise division in FlexRound.
% This section begins by introducing the notations used throughout the paper. We then provide an overview of the concept and design of FlexRound, a per-tensor uniform post-training quantization (PTQ) reconstruction method. We also delve into the advantages of utilizing the principle of element-wise division in FlexRound.
%In this section, we first present the notations used in the paper, describe the concept and design of FlexRound for per-tensor uniform post-training quantization (PTQ) reconstruction, and then, scrutinize \textcolor{blue}{how FlexRound can benefit from the principle of element-wise division.} 
% leverage the importance of a pre-trained weight.

\subsection{Preliminaries}\label{subsec:preliminaries}

\paragraph{Notations} A scalar, a vector, and a matrix (or a tensor) are expressed as a non-bold letter, a small bold letter, and a capital bold letter (e.g. $s$, $\vs$ and $\mS$) respectively. $\widehat{\mW}$ indicates the quantized counterpart of $\mW$. 
The input to a 2D convolution or a linear layer is represented as $\mX$ if all previous layers are intact, or as $\widetilde{\mX}$ if all previous layers are quantized. 
The entries of a matrix $\mA$ are denoted as $A_{(i, j)}$, while the entries of a 4-dimensional tensor $\mA$ are denoted as $A_{(i, j, k, l)}$. We let $\odot$ and $\mathbin{/}$ indicate element-wise product and element-wise division, respectively, similar to the broadcasting process in Python Numpy. 
% For example, $\mY = \mW \odot \vx$ and $\mC = {\mA \over \mB}$ imply that $Y_{(i, j)} = W_{(i, j)} x_{(i)}$ and $C_{(i, j)} = {A_{(i, j)} \over B_{(i, j)}}$ for every $i$ and $j$. 
$\lfloor \cdot \rceil$ and $\lfloor \cdot \rfloor$ express the rounding function and the floor function. $|| \cdot ||_F$ represents the Frobenius norm.

\paragraph{PTQ Background} The conventional uniform PTQ approach is to quantize pre-trained weights $\mW$ to be $\widehat{\mW} = s_1 \Big\lfloor {\mW \over s_1} \Big\rceil$ via rounding-to-nearest, where a quantization grid size $s_1$ $\in \mathbb{R}_{> 0}$ is set to minimize $\|\mW - \widehat{\mW}\|^2_F$, but the minimization of the quantization error in the parameter space is not equivalent to that of the final task loss. As \citet{li2021brecq} proves that the loss degradation resulting from quantization can be approximated as the quadratic form of the network output and its Hessian matrix, several studies have strove to minimize $\|\mW\mX - \widehat{\mW}\widetilde{\mX}\|^2_F$ layer-by-layer or block-by-block with respect to continuous variables $\mV$ with a small amount of data, where $\widehat{\mW}$ is either $s_1 (\lfloor {\mW \over s_1} \rfloor + h(\mV))$ with a certain function $h(\cdot)$ \citep{nagel2020adaround} or $s_1 \Big\lfloor {\mW + \mV \over s_1} \Big\rceil$ \citep{hubara2021adaquant}. However, all these rounding mechanisms are founded on element-wise addition. % to learn where to quantize pre-trained weights

\subsection{FlexRound}\label{subsec:FlexRound}

% \begin{figure}[t]
% 	\centering
% 	\subfigure{\includegraphics[width=\linewidth]{figures/figure1.pdf}}
% 	\caption{Illustration of FlexRound.}
% 	\label{fig:FlexRound}
% \end{figure}

Unlike prior works based on element-wise addition, we exploit element-wise division for quantizing pre-trained weights. We can formulate our proposed weight-rounding scheme based on element-wise division as follows:
\begin{align}
    \widehat{\mW} = s_1 \Big\lfloor {\mW \over \mS} \Big\rceil, \label{eq:flexround}
\end{align}
where $\mS$ is the division factor for $\mW$ whose shape is equal to that of $\mW$ while all entries of $\mS$ as well as $s_1$ are positive and learnable. Similarly to preceding studies, both $s_1$ and $\mS$ are updated as an attempt to minimize $\|\mW\mX - \widehat{\mW}\widetilde{\mX}\|^2_F$.

Eq.~\ref{eq:flexround} indicates that the basic formula of FlexRound supports per-tensor uniform PTQ. Although FlexRound can also adopt per-channel weight quantization by simply replacing a scalar $s_1$ with a vector $\vs_1$, as we show later, per-tensor uniform PTQ (via FlexRound) % is
can be sufficient to achieve the performance of a full-precision model. 
% Therefore, we set a single quantization grid size $s_1$ for each layer. Note that a per-tensor quantization scheme can improve the efficiency of the quantized model during inference, such as through the use of integer matrix-to-matrix multiplication API/function calls \citep{migacz20178}. 
Therefore, we focus on the per-tensor uniform PTQ reconstruction unless otherwise specified. The overall procedure of FlexRound in a per-tenor uniform PTQ setting is described in Figure~\ref{fig:FlexRound}.
% Note, however, that FlexRound can be easily used for per-channel weight quantization.
%  like most of the uniform PTQ works have done
% to achieve comparable performance to a full-precision model, 
% most of the uniform PTQ works rely on per-channel weight quantization rather than per-layer weight quantization to achieve comparable performance to a full-precision model. However, In order to show that the performance of a full-precision model can be maintained even after per-tensor uniform PTQ, we set a single scaling factor $s_1$ for each layer.

% \textcolor{red}{(ChatGPT) Eq. 1 indicates that the basic formula of FlexRound supports per-tensor uniform PTQ. Although FlexRound can also adopt per-channel weight quantization by simply replacing a scalar s1 with a vector s1, as we will show later, per-tensor uniform PTQ (via FlexRound) is sufficient to achieve the performance of a full-precision model. Therefore, we set a single quantization grid size s1 for each layer. Note that a per-tensor quantization scheme can improve the efficiency of the quantized model during inference, such as through the use of integer matrix-to-matrix multiplication API/function calls (Migacz, 2017). Therefore, we will focus on per-tensor uniform PTQ reconstruction in the rest of this analysis. The overall procedure of FlexRound is described in Figure 1.}

\begin{figure}
    % \vskip -10pt
    % \vskip -1pt
    \centering
    \subfigure{\includegraphics[scale=0.24]{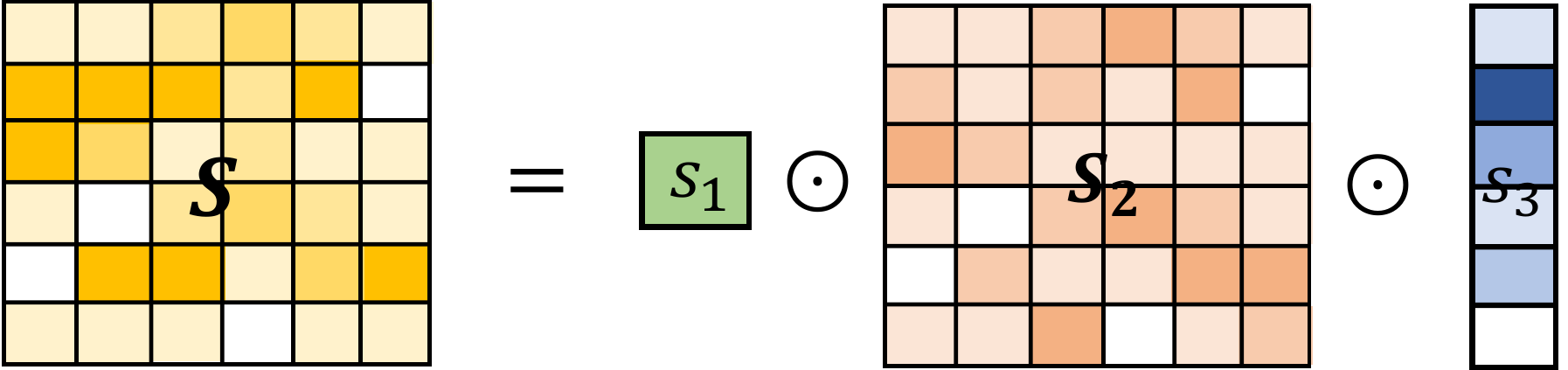}}
    % \vskip -8pt
    \caption{Formation of $\mS$ in Eq.~\ref{eq:flexround} for a linear layer $\mW$. $s_1$ is a common quantization grid size across a layer, $\mS_2$ is the matrix scaling $\mW$, and $\vs_3$ is an additional vector supporting $\mS_2$ to account for the variation of output channel's statistics in $\mW$. As a result, $\mS = s_1 \odot \mS_2 \odot \vs_3$ is the division factor for a linear layer $\mW$. }
    % \vskip -12pt
    \label{fig:formation}
    \vskip -0.15in
\end{figure}

Let us discuss how to design $\mS$ in detail. We first start formulating $\mS$ as $\mS = s_1 \odot \mS_2$, where $\mS_2$ is the matrix or tensor scaling $\mW$ whose shape is equal to that of $\mW$ while every element of $\mS_2$ is positive and learnable. When $\mS = s_1 \odot \mS_2$, Eq.~\ref{eq:flexround} is enough to perform well compared to existing weight-rounding schemes based on element-wise addition in a per-tensor uniform PTQ setting, as we show later. However, to further improve the performance of a new weight-rounding scheme based on element-wise division, % in a per-tensor uniform PTQ setting, 
we complement $\mS_2$ as follows. For a linear layer $\mW \in \mathbb{R}^{C_{out} \times C_{in}}$, $\mS_2$ is complemented with an additional learnable tensor $\vs_3 \in \mathbb{R}_{> 0}^{C_{out} \times 1}$. Motivated from a wide acknowledgement that the statistics of output channels can vary greatly \citep{nagel2019data, lou2020autoq}, we take into account the variation of output channel's statistics by supplementing $\mS_2$ with $\vs_3$. 
For a 2D convolution $\mW \in \mathbb{R}^{C_{out} \times C_{in} \times H \times W}$, in particular, $\mS_2$ is complemented with two additional learnable tensors $\vs_3 \in \mathbb{R}_{> 0}^{C_{out} \times 1 \times 1 \times 1}$ and $\vs_4 \in \mathbb{R}_{> 0}^{1 \times C_{in} \times 1 \times 1}$. Hence, $\mS$ is formulated as $s_1 \odot \mS_2 \odot \vs_3$ (as illustrated in Figure~\ref{fig:formation}) for a linear layer, or as $s_1 \odot \mS_2 \odot \vs_3 \odot \vs_4$ for a 2D convolution so that Eq.~\ref{eq:flexround} is transformed into
\begin{equation}
    \widehat{\mW} = 
    \begin{cases}
    s_1 \Big\lfloor {\mW \over {s_1 \odot \mS_2 \odot \vs_3}} \Big\rceil \quad\,\, \text{ for a linear layer} \\
    s_1 \Big\lfloor {\mW \over {s_1 \odot \mS_2 \odot \vs_3 \odot \vs_4}} \Big\rceil \,\text{for a 2D convolution} 
    \end{cases}\label{eq:flexround2}.
\end{equation}
% \begin{equation}
%     \widehat{\mW} =  
%     \begin{cases}
%     s_1 \Big\lfloor {\mW \over {s_1 \odot \mS_2 \odot \vs_3}} \Big\rceil \quad\,\, \text{ if } \mW \text{ is a linear layer} \\
%     s_1 \Big\lfloor {\mW \over {s_1 \odot \mS_2 \odot \vs_3 \odot \vs_4}} \Big\rceil \,\text{if } \mW  \text{ is a 2D convolution}  
% \end{cases}\label{eq:flexround2}
% \end{equation}
We refer to Eq.~\ref{eq:flexround2} as `\emph{FlexRound.}' Here, every element of $\mS_2$, $\vs_3$, and $\vs_4$ is initialized to 1 in order to facilitate learning from the traditional rounding-to-nearest method, namely, $s_1 \Big\lfloor {\mW \over s_1} \Big\rceil$. All parameters ($s_1$, $\mS_2$, $\vs_3$, and $\vs_4$) are updated to minimize $\|\mW\mX - \widehat{\mW}\widetilde{\mX}\|^2_F$ subject to the constraint that all entries of $s_1$, $\mS_2$, $\vs_3$, and $\vs_4$ are positive.

\begin{figure*}
    \vskip -0.05in
    % \vskip -12pt
    \centering
    \subfigure[MobileNetV2]
    {
        \includegraphics[width=0.9\linewidth]{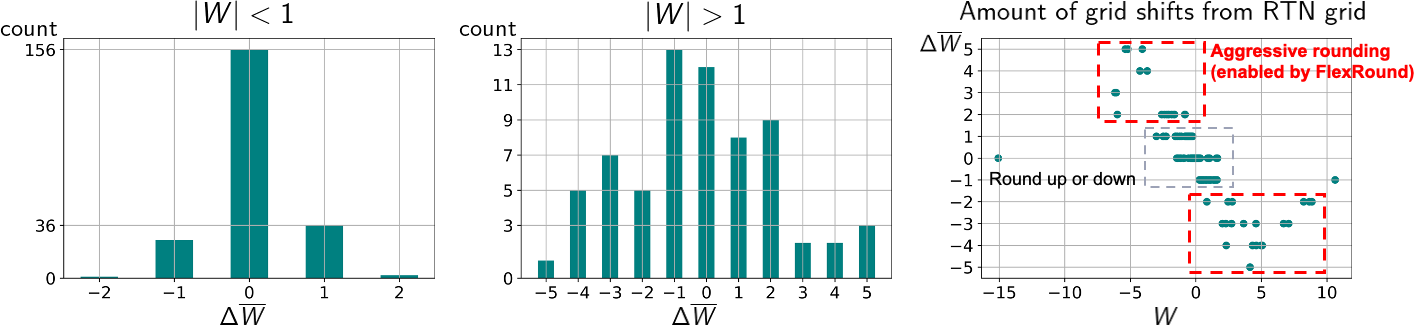}
        \label{fig:mobilenet}
    }
    \\
    \vskip -0.01in
    \subfigure[ResNet-18]
    {
        \includegraphics[width=0.9\linewidth]{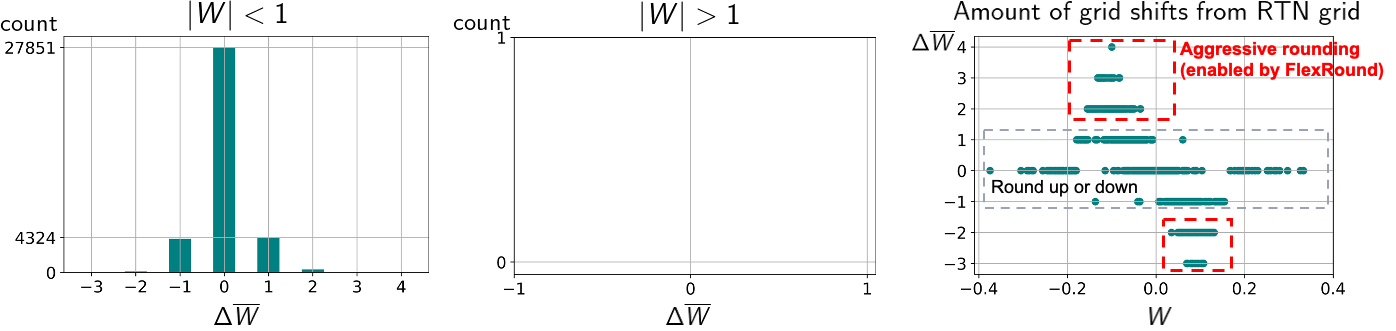}
        \label{fig:resnet}
    }
    % \vskip -5pt
    \vskip -0.1in
    \caption{Weight updates through FlexRound of the first 2D convolution in the first block of (a) MobileNetV2 and (b) ResNet-18, after quantizing pre-trained weights to $4$-bit (via FlexRound) while activations are kept in full-precision.} 
    \label{fig:histogram}
    % \vskip -10pt
    \vskip -0.05in
\end{figure*}

% \begin{figure*}
%     % \vskip -12pt
%     \vskip -0.1in
%     \centering
%     \subfigure[] % Example of FlexRound scheme
%     {
%         \includegraphics[width=0.3\linewidth]{figures/Figure_B.png}
%         \label{fig4:a}
%     }
%     \subfigure[]
%     {
%         \includegraphics[width=0.3\linewidth]{figures/Figure_A.png}
%         \label{fig4:b}
%     }
%     % \vskip -5pt
%     \vskip -0.1in
%     \caption{Amount of grid shifts from the grids obtainable from RTN through FlexRound (a) in the second 2D convolution of the sixth block of MobileNetV2 when only weights are quantized to $4$-bit and (b) in the query projection of the first self-attention layer of $\text{BERT}_{\text{BASE}}$ fine-tuned on the MRPC dataset when quantizing both weights and input activations of self-attention and feed-forward layers to $8$-bit. Unlike the right side of Figure \ref{fig:histogram}, weights of large magnitude are quantized with similar flexibility to those of moderate magnitude.}
%     \label{fig:FlexRound}
%     % \vskip -10pt
%     \vskip -0.1in
% \end{figure*}

\begin{figure}
    \vskip -0.1in
    % \vskip -10pt
    % \vskip 2pt
    \centering
    \subfigure{\includegraphics[width=0.76\linewidth]{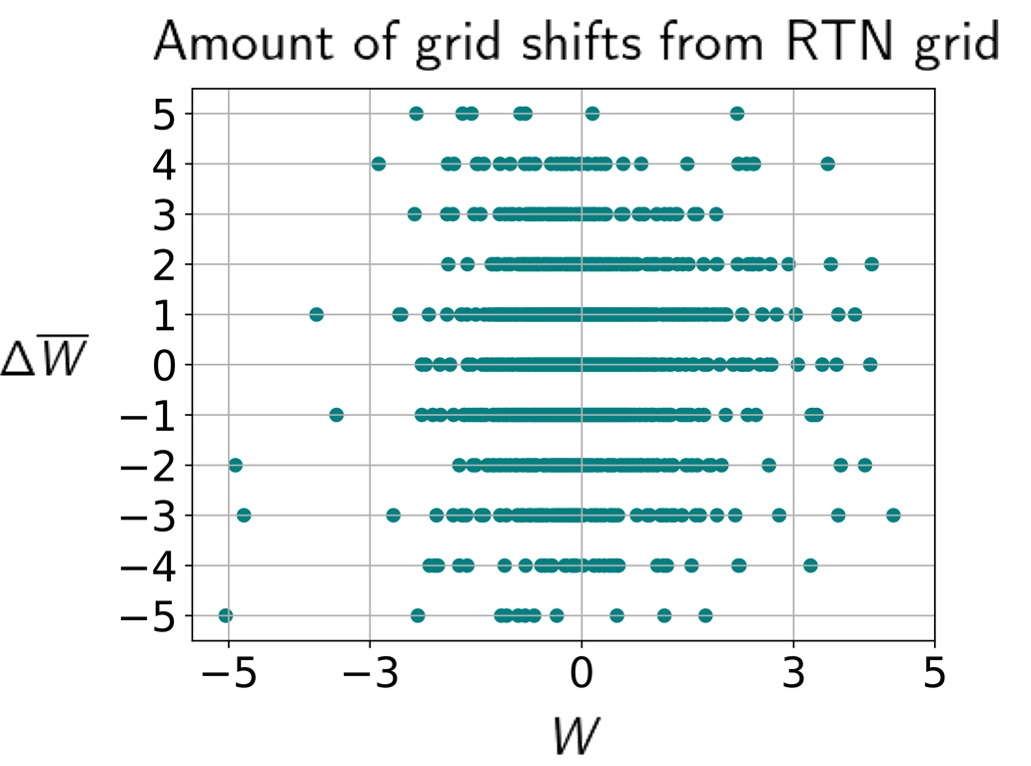}}
    \vskip -0.1in
    \caption{Amount of grid shifts from the grids obtainable from RTN in the second 2D convolution of the sixth block of MobileNetV2 when only weights are quantized to $4$-bit via FlexRound. Unlike the right side of Figure \ref{fig:histogram}, weights of large magnitude are quantized with similar flexibility to those of moderate magnitude.}
    % \vskip -12pt
    \label{fig:B}
    \vskip -0.2in
\end{figure}

In Eq.~\ref{eq:flexround2}, element-wise division serves a similar purpose as element-wise addition in creating a more effective rounding scheme than rounding-to-nearest. By implementing such a new rounding policy through element-wise division, we can make $s_1$, $\mS_2$, $\vs_3$, and $\vs_4$ all learnable. This allows FlexRound to learn a common quantization grid size (i.e., $s_1$) jointly with the rounding process (e.g., $\mS_2 \odot \vs_3$ or $\mS_2 \odot \vs_3 \odot \vs_4$ in FlexRound). Furthermore, the reciprocal rule of derivatives induced by element-wise division enables FlexRound to leverage pre-trained weights when learning the corresponding scales, as demonstrated both theoretically and empirically by the following proposition.

\begin{proposition}\label{prop}
    Let $\mathcal{L}$ be the reconstruction error computed from Eq.~\ref{eq:flexround2} and $\mS'$ be the matrix (or tensor) scaling pre-trained weights $\mW$ in Eq.~\ref{eq:flexround2}, i.e., $\mS' = \mS_2 \odot \vs_3$ (or $\mS_2 \odot \vs_3 \odot \vs_4$). Then, the gradient of $\mathcal{L}$ with respect to an entry of $\mS'$, ${{\partial\mathcal{L}} \over {\partial S'_{(i, j)}}}$ (or ${{\partial\mathcal{L}} \over {\partial S'_{(i, j, k, l)}}}$) is proportional to its corresponding pre-trained weight, $W_{(i, j)}$ (or $W_{(i, j, k, l)}$), when using the straight-through estimator \citep{Bengio2013ste}.
    % \textcolor{red}{The reconstruction error, $\mathcal{L}$, is computed from Eq.~\ref{eq:flexround2}. The matrix or tensor, $\mS'$, scales pre-trained weights $\mW$, where $\mS' = \mS_2 \odot \vs_3$ (or $\mS_2 \odot \vs_3 \odot \vs_4$) using the straight-through estimator \citep{Bengio2013ste}. The gradient of $\mathcal{L}$ with respect to an element of $\mS'$, $\left|{{\partial\mathcal{L}} \over {\partial S'_{(i, j)}}}\right|$ (or $\left|{{\partial\mathcal{L}} \over {\partial S'_{(i, j, k, l)}}}\right|$) is directly proportional to its corresponding pre-trained weight, $\left|W_{(i, j)}\right|$ (or $\left|W_{(i, j, k, l)}\right|$).}
\end{proposition}

Proposition~\ref{prop} implies that, for a linear layer, an element $S'_{(i, j)}$ is (partially) affected by $W_{(i, j)}$ so that $\overline{W}_{(i, j)} = \Big\lfloor{W_{(i, j)} \over {s_1 \odot S'_{(i, j)}}}\Big\rceil$ can also be updated and influenced by $W_{(i, j)}$.
% \textcolor{red}{Proposition~\ref{prop} implies that, for a linear layer, the element $S'_{(i, j)}$ is (partially) affected by $W_{(i, j)}$ such that $\overline{W}_{(i, j)} = \Big\lfloor{W_{(i, j)} \over {s_1 \odot S'_{(i, j)}}}\Big\rceil$ can also be updated and influenced by $W_{(i, j)}$.}
% Using the straight-through estimator \citep{Bengio2013ste}, for every $i$ and $j$, $\left|{{\partial\mathcal{L}} \over {\partial S'_{(i, j)}}}\right|$ is directly proportional to $\left|W_{(i, j)}{{\partial\mathcal{L}} \over {\partial \widehat{W}_{(i, j)}}}\right|$, which implies that $S'_{(i, j)}$ is (partially) affected by $W_{(i, j)}$. As a result, $\overline{W}_{(i, j)} = \textcolor{blue}{\Big\lfloor}{W_{(i, j)} \over {s_1 \odot S'_{(i, j)}}}\textcolor{blue}{\Big\rceil}$ can also be updated and influenced by $W_{(i, j)}$ as well. 
In other words, as the magnitude of a pre-trained weight $W_{(i, j)}$ is larger, the chance of $\overline{W}_{(i, j)}$ receiving a larger update during the PTQ reconstruction becomes higher. The magnitude of a weight can be regarded as a metric to measure the importance of a weight for pruning unimportant weights \citep{han2015learning}. Consequently, weights of larger magnitude play a more important role than those of smaller magnitude \citep{han2016deep}.
%if the goal is to enhance model accuracy after quantization, 
To reduce the performance gap between a full-precision pre-trained model and its quantized version, it would be reasonable to relax the constraint on quantizing pre-trained weights of large magnitude (i.e., potentially important pre-trained weights) by allowing them to have higher chances of being quantized to one of not just the two closest quantization grids but also more distant ones than those of smaller magnitude.
% it would be reasonable to have \textcolor{blue}{weights of smaller magnitude (that is, less important weights)} rounded either up or down only while allowing \textcolor{blue}{weights of larger magnitude (i.e., more important weights)} to be quantized to one of the two closest quantization grids or more.
The above implication is also identically applicable to a 2D convolution.

% \textcolor{red}{(ChatGPT) Proposition 3.1 implies that, for a linear layer, an element S(i,j) is (partially) affected by W(i,j) such that WW(i,j) can also be updated and influenced by W(i,j). In other words, as the magnitude of a pre-trained weight W(i,j) increases, the chance of WW(i,j) receiving a larger update during the PTQ reconstruction also increases. The magnitude of a weight can also be used as a metric to measure the importance of a weight for pruning unimportant weights (Han et al., 2015). Therefore, weights with larger magnitude play a more important role than those with smaller magnitude (Han et al., 2016). To reduce the performance gap between the model before and after quantization, it would be reasonable to relax the constraint on quantizing pre-trained weights of large magnitude (i.e., important pre-trained weights) by allowing them to have more possibilities to be quantized to one of not just the two closest quantization grids but also more ones than those of smaller magnitude. The above implication is also applicable to the case of a 2D convolution.}

Figure~\ref{fig:histogram} shows the amount of weight updates via FlexRound for MobileNetV2 and ResNet-18. 
On the left side and the center side of Figure~\ref{fig:histogram}, histograms describe the change of $\overline{W}_{(i, j, k, l)}$ grouped for small pre-trained weights ($|W| < 1$, left) and large pre-trained weights ($|W| > 1$, center). 
On the right side, scatter plots show the amount of grid shifts from the grids obtainable from rounding-to-nearest (RTN).
We note that MobileNetV2 and ResNet-18 are quantized distinctively due to FlexRound.
For example, in the case of MobileNetV2 as in Figure~\ref{fig:mobilenet}, the change of $\overline{W}_{(i, j, k, l)}$ attained by minimizing $\mathcal{L}$ is more aggressive (i.e., rounding can be deviated from more than one-step up or one-step down) when the absolute value of $W_{(i, j, k, l)}$ is larger than one, which means that FlexRound more flexibly quantizes pre-trained weights of large magnitude as illustrated in red dotted squares in Figure~\ref{fig:mobilenet}. The amount of aggressively rounded weights in the first 2D convolution of the first block of MobileNetV2 is around $12.8\%$ of the total. For ResNet-18, however, there are no pre-trained weights whose magnitudes are larger than one. Thus, most pre-trained weights are rounded either up or down as seen in Figure~\ref{fig:resnet} (e.g., only about $1.5\%$ weights are rounded aggressively in the first 2D convolution of the first block of ResNet-18). Different rounding results of AdaRound, AdaQuant, and FlexRound are visually compared in Appendix~\ref{appendix:comparison}. %, which leads FlexRound to perform slightly better than previous methods as evidenced in the third column of Table \ref{tab:imagenet_w} and \ref{tab:imagenet_wa}.

\begin{figure}
    % \vskip -10pt
    % \vskip 2pt
    \vskip -0.1in
    \centering
    \subfigure{\includegraphics[width=0.76\linewidth]{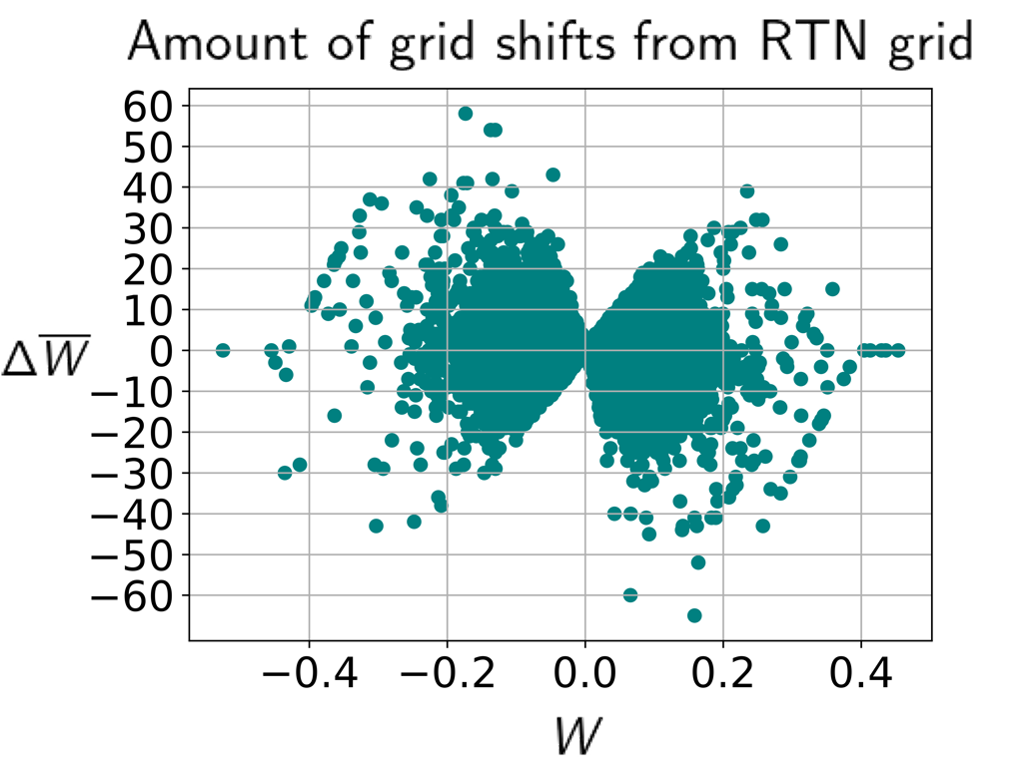}}
    \vskip -0.1in
    \caption{Number of grid shifts from the grids attainable from RTN in the query projection of the first self-attention layer of $\text{BERT}_{\text{BASE}}$ fine-tuned on the MRPC dataset when quantizing both weights and input activations of self-attention and feed-forward layers to $8$-bit via FlexRound. FlexRound can provide up to about $60$ grid shifts from the grids obtainable from RTN.}
    % \vskip -12pt
    \label{fig:A}
    \vskip -0.2in
\end{figure}

\begin{table*}[t]
\vskip -0.1in
% \vskip -12pt
\caption{Top-1/Top-5 accuracy (\%) on ImageNet when only weights are quantized to $4$-bit. ``B $+$ X" denotes the implementation of X in the setting of BRECQ. The $s_1$ column indicates whether $s_1$ is fixed or can be learned during the PTQ reconstruction. The $\mS_2$ and $\vs_3, \vs_4$ columns represent the presence (O) or absence (X) of each in FlexRound, respectively. For instance, the formula for FlexRound (Ours) and Ablation Study 1 is Eq.~\ref{eq:flexround2}, and that for Ablation Study 2 is $\widehat{\mW} = s_1 \lfloor \mW / s_1 \odot \mS_2 \rceil$.} % of ResNet-18, ResNet-50, and MobileNetV2 
\label{tab:ablation}
\begin{center}
\small
\begin{tabular}{lcccccc}
\toprule
\makecell{Method} & $s_1$ & $\mS_2$ & $\vs_3, \vs_4$ & \makecell{ResNet-18} & \makecell{ResNet-50} & \makecell{MobileNetV2} \\
\midrule
Full-precision & N/A & N/A & N/A & $71.00 / 89.97$ & $76.63 / 93.04$ & $72.62 / 90.67$\\
\midrule
B + AdaQuant & Learnable & N/A & N/A & $67.50 / 87.75$ & $72.79 / 90.77$ & $15.17 / 32.89$  \\
B + AdaRound & Fixed & N/A & N/A & $70.18 / 89.38$ & $75.86 / 92.62$ & $69.46 / 88.85$ \\
\midrule
B + FlexRound (Ours) & Learnable & O & O & $\mathbf{70.28} / \mathbf{89.44}$ & $\mathbf{75.95} / \mathbf{92.68}$ & $\mathbf{70.82} / \mathbf{89.67}$ \\ 
$\rightarrow$ Ablation Study 1 & Fixed & O & O & $70.09/ 89.43$ & $75.88 / 92.61$ & $69.47 / 88.85$ \\ %  with fixed $s_1$ 
$\rightarrow$ Ablation Study 2 & Learnable & O & X & $70.22/ 89.45$ & $75.92 / 92.63$ & $70.51 / 89.49$ \\ %  with \textcolor{blue}{only $s_1$ and $\mS_2$} 
\bottomrule
\end{tabular}
\end{center}
% \vskip -10pt
\vskip -0.2in
\end{table*}

Even if FlexRound takes into account the magnitude of pre-trained weights when updating their corresponding scales, one might question that FlexRound seems to quantize pre-trained weights of moderate magnitude more flexibly than those of large magnitude as seen in the right side of Figure~\ref{fig:histogram}. Our aim with FlexRound is to emphasize that pre-trained weights with relatively larger magnitude are more likely to be quantized with higher flexibility compared to those with relatively smaller magnitude. As explained in Appendix~\ref{appendix:proof}, $\left| {{\partial\mathcal{L}} \over {\partial S'_{(i, j)}}} \right|$ is directly proportional to $\left| W_{(i, j)} {{\partial\mathcal{L}} \over {\partial \widehat{W}_{(i, j)}}} \right|$. No matter how large the magnitude of $W_{(i, j)}$ is, if $\left| {{\partial\mathcal{L}} \over {\partial \widehat{W}_{(i, j)}}} \right|$ is close to zero, $\left| {{\partial\mathcal{L}} \over {\partial S'_{(i, j)}}} \right|$ would be also zero. In this sense, pre-trained weights of large magnitude can be quantized to the grids obtainable from RTN. If $\left| {{\partial\mathcal{L}} \over {\partial \widehat{W}_{(i, j)}}} \right|$ is (significantly) larger than zero, pre-trained weights of large magnitude can be quantized to the grids far from two nearest ones as seen in Figure~\ref{fig:B}. In short, while the magnitude of pre-trained weights influences the updates to their corresponding scales in FlexRound, it does not necessarily imply that larger weights must be quantized more flexibly than smaller ones.

% \textcolor{blue}{Although FlexRound takes into account the magnitude of pre-trained weights when updating their corresponding scales, one might question that FlexRound seems to quantize pre-trained weights of moderate magnitude more flexibly than those of large magnitude. Our aim with FlexRound is to emphasize that pre-trained weights with relatively larger magnitudes are more likely to be quantized with higher flexibility compared to those with relatively smaller magnitudes. $\left| {{\partial\mathcal{L}} \over {\partial S'_{(i, j)}}} \right|$ is directly proportional to $\left| W_{(i, j)} {{\partial\mathcal{L}} \over {\partial \widehat{W}_{(i, j)}}} \right|$ as seen in Appendix \ref{appendix:proof}. No matter how large the magnitude of $W_{(i, j)}$ is, if $\left| {{\partial\mathcal{L}} \over {\partial \widehat{W}_{(i, j)}}} \right|$ is close to zero, then $\left| {{\partial\mathcal{L}} \over {\partial S'_{(i, j)}}} \right|$ would be also zero. In this sense, as seen in the right side of Figure \ref{fig:histogram}, pre-trained weights of large magnitude can be quantized to the grids obtainable from RTN. If $\left| {{\partial\mathcal{L}} \over {\partial \widehat{W}_{(i, j)}}} \right|$ is (significantly) larger than zero, then pre-trained weights of large magnitude can be quantized to the grids far from two nearest ones as shown in Figure \ref{fig:B}. In short, while the magnitude of pre-trained weights influences the updates to their corresponding scales in FlexRound, it does not necessarily imply that larger weights must be quantized more flexibly than smaller ones.}

Note that FlexRound can quantize weights more flexibly as the bit-width increases. Comparing the right side of Figure~\ref{fig:histogram} with Figure~\ref{fig:A}, FlexRound can provide more grid shifts from the grids obtainable from RTN as a higher bit-width is used. Unlike AdaRound that must round weights either up or down regardless of the number of bits used, FlexRound enables more flexible weight quantization as the bit-width increases, thus being better suited for quantizing models that require higher bit-widths (e.g., LLMs) than AdaRound. % FlexRound allows for more flexible weight quantization as higher bit-widths are utilized. 

\begin{table*}[t]
\vskip -0.1in
\caption{Top-1/Top-5 accuracy (\%) on ImageNet with only weights quantized. ``B $+$ X" is the implementation of X in the BRECQ's setup.} % of ResNet-18, ResNet-50, and MobileNetV2 
\label{tab:imagenet_w_brecq}
\begin{center}
\small
\begin{tabular}{lcccc}
\toprule
Method & \# Bits (W/A) & ResNet-18 & ResNet-50 & MobileNetV2 \\
\midrule
Full-precision & $32 / 32$ & $71.00 / 89.97$ & $76.63 / 93.04$ & $72.62 / 90.67$\\
\midrule
B + AdaQuant & $4 / 32$ & $67.50 / 87.75$ & $72.79 / 90.77$ & $15.17 / 32.89$  \\
B + AdaRound & $4 / 32$ & $70.18 / 89.38$ & $75.86 / 92.62$ & $69.46 / 88.85$ \\
B + FlexRound (Ours) & $4 / 32$ & $\mathbf{70.28} / \mathbf{89.44}$ & $\mathbf{75.95} / \mathbf{92.68}$ & $\mathbf{70.82} / \mathbf{89.67}$ \\ 
\midrule
B + AdaQuant & $3 / 32$ & $57.09 / 80.82$ & $52.13 / 75.22$ & $0.20 / 0.79$  \\
B + AdaRound & $3 / 32$ & $\mathbf{68.79} / \mathbf{88.62}$ & $74.31 / 91.81$ & $62.51 / 84.52$ \\
B + FlexRound (Ours)& $3 / 32$ & $68.65 / 88.54$ & $\mathbf{74.38} / \mathbf{91.81}$ & $\mathbf{66.87} / \mathbf{87.56}$ \\
\midrule
B + AdaQuant & $2 / 32$ & $0.23 / 0.92$ & $0.10 / 0.50$ & $0.10 / 0.50$  \\
B + AdaRound & $2 / 32$ & $61.99 / 84.81$ & $48.47 / 77.09$ & $39.57 / 66.18$ \\ % & $63.07 / 85.09$ & $68.33 / 89.23$ & $32.59 / 60.05$ \\ % Note that ResNet-50 2-bit \emph{per-channel asymmetric} quantization: $72.08 / 90.85$ and \emph{per-channel symmetric} quantization: $69.43 / 89.60$
B + FlexRound (Ours)& $2 / 32$ & $\mathbf{62.57} / \mathbf{84.84}$ & $\mathbf{63.67} / \mathbf{85.72}$ & $\mathbf{46.04} / \mathbf{72.48}$ \\
\bottomrule
\end{tabular}
\end{center}
% \vskip -10pt
\vskip -0.15in
\end{table*}

\begin{table*}[t]
% \vskip -12pt
\vskip -0.1in
\caption{Top-1/Top-5 accuracy (\%) on ImageNet when both weights and activations are quantized. ``B $+$ X" and ``Q $+$ Y" represent the implementation of X in the BRECQ's setting and that of Y in the QDrop's setting, respectively.} % of ResNet-18, ResNet-50, and MobileNetV2 
\label{tab:imagenet_wa_brecq}
\begin{center}
\small
\begin{tabular}{lcccc}
\toprule
\makecell{Method} & \makecell{\# Bits (W/A)} & \makecell{ResNet-18} & \makecell{ResNet-50} & \makecell{MobileNetV2} \\
\midrule
Full-precision & $32 / 32$ & $71.00 / 89.97$ & $76.63 / 93.04$ & $72.62 / 90.67$\\
\midrule
B + AdaRound & $4 / 4$ & $69.18 / 88.85$ & $74.44 / 91.80$ & $61.05 / 83.30$  \\
B + FlexRound (Ours)& $4 / 4$ & $\mathbf{69.32} / \mathbf{88.83}$ & $74.56 / 91.87$ & $63.74 / 85.01$ \\ 
Q + AdaRound & $4 / 4$ & $69.20 / 88.96$ & $74.90 / 92.15$ & $65.42 / 86.23$  \\
Q + FlexRound (Ours)& $4 / 4$ & $69.26 / 88.81$ & $\mathbf{75.08} / \mathbf{92.20}$ & $\mathbf{66.66} / \mathbf{87.21}$ \\
\midrule
B + AdaRound & $3 / 3$ & $64.83 / 86.12$ & $67.01 / 87.28$ & $3.74 / 11.54$  \\
B + FlexRound (Ours)& $3 / 3$ & $64.99 / 85.93$ & $68.29 / 87.89$ & $25.43 / 48.28$ \\ 
Q + AdaRound & $3 / 3$ & $\mathbf{65.71} / \mathbf{86.96}$ & $70.49 / 89.93$ & $39.86 / 66.00$  \\
Q + FlexRound (Ours)& $3 / 3$ & $65.43 / 86.60$ & $\mathbf{70.74} / \mathbf{89.78}$ & $\mathbf{51.49} / \mathbf{76.90}$ \\
\bottomrule
\end{tabular}
\end{center}
% \vskip -10pt
\vskip -0.15in
\end{table*}

\section{Experiments}\label{sec:experiments}
In this section, % we present experimental results for benchmark datasets and models in computer vision and natural language processing tasks.
we first empirically confirm the importance of learning a quantization grid size $s_1$ jointly with the rounding process and the distinct contribution of additional tensors $\vs_3$ and $\vs_4$ to FlexRound. % in a per-tensor uniform PTQ setting.
% We first empirically confirm that learning a quantization grid size $s_1$ jointly with rounding is crucial for designing a better rounding scheme than existing state-of-the-art rounding schemes and that additional tensors $\vs_3$ and $\vs_4$ introduced in Section~\ref{subsec:FlexRound} implement distinct contributions in a per-tensor uniform post-training quantization (PTQ) setting.
% to verify that FlexRound can achieve similar performance to a full-precision model for the above tasks in a per-tensor uniform PTQ setting,
% \textcolor{blue}{to verify that FlexRound can perform similarly to a full-precision model even in a per-tensor uniform PTQ setting,} 
Then, we compare the performance of FlexRound with that of the state-of-the-art PTQ methods in a per-tensor uniform PTQ setting in the following cases: image classification on ImageNet \citep{russakovsky2015imagenet} with ResNet \citep{he2016deep} and MobileNetV2 \citep{sandler2018mobilenetv2} (Section~\ref{subsec:imagenet}), natural language understanding (NLU) on GLUE \citep{wang2018glue} with BERT \citep{devlin2018bert} and GPT-Neo \citep{gpt-neo} (Section~\ref{subsec:nlu}), natural language generation (NLG) on WikiText2 \citep{merity2016pointer} and Penn Treebank (PTB) \citep{marcus-etal-1993-building} with GPT-Neo and OPT \citep{zhang2022opt}, and NLG on WebNLG \citep{gardent2017webnlg} with GPT-2 \citep{radford2019gpt2} (Section~\ref{subsec:nlg}). 
% Throughout the comprehensive experiments, we verify that FlexRound can achieve competitive performance with a full-precision model for the above tasks even in a per-tensor uniform PTQ setting. Not only that,
Finally, we validate that large language models (LLMs) can be quantized with only a marginal impact on accuracy compared to half-precision baselines by block-wise output reconstruction, without assuming that the activation outliers would occur in a certain pattern. We study LLaMA \citep{touvron2023llama} by adopting per-channel weight quantization and per-tensor activation quantization for six common sense reasoning benchmarks: BoolQ \citep{clark2019boolq}, PIQA \citep{bisk2020piqa}, HellaSwag \citep{zellers2019hellaswag}, WinoGrande \citep{sakaguchi2021winogrande}, ARC easy and challenge \citep{clark2018arc}, and OpenBookQA \citep{mihaylov2018obqa}, and the causal language modeling task on WikiText2 (Section~\ref{subsec:nlg}).

For brevity, we let ``B + X'' and ``Q + X'' indicate that a certain rounding scheme `X' is performed in the experimental setup described in BRECQ \citep{li2021brecq} or QDrop \citep{wei2022qdrop}, respectively (an experimental setup includes the definition of a block unit for reconstruction error minimization or how much the probability of dropping the quantization of activations is). As introduced in BRECQ and QDrop, we also use the LSQ technique \citep{esser2020learned} when updating an activation step size for activation quantization. 
% Throughout our comprehensive experiments, we verify that FlexRound can achieve competitive performance with a full-precision model for the above tasks even in a \emph{per-tensor} uniform PTQ setting. % , which has not been introduced previously. 
All experimental results are conducted by our own implementation based on open-source codes.

\begin{table*}[t]
\vskip -0.1in
\caption{Performance on GLUE. For evaluation metrics, matched and mismatched accuracies are reported for MNLI, F1 score and accuracy are reported for QQP, and accuracy is reported for MRPC. ``Q $+$ X" implies the implementation of X in the QDrop's setting. Both weights and input activations of attention and feed-forward sub-layers are quantized to $8$-bit in a per-tensor asymmetric scheme.} 
\label{tab:glue}
\begin{center}
\small
% \resizebox{0.9\linewidth}{!}{
\begin{tabular}{clccccc}
\toprule
Dataset & \makecell{Method} & $\text{BERT}_{\text{BASE}}$ & $\text{BERT}_{\text{LARGE}}$ & $\text{GPT-Neo}_{125\text{M}}$ & $\text{GPT-Neo}_{1.3\text{B}}$ & $\text{GPT-Neo}_{2.7\text{B}}$ \\
% Method & \makecell{\# Bits \\ (W./A.)} & \makecell{MNLI \\ (acc. m/mm)} & \makecell{QQP \\ (F1/acc.)} & \makecell{QNLI \\ (acc.)} & \makecell{SST-2 \\ (acc.)} & \makecell{CoLA \\ (Matthews corr.)} & \makecell{STS-B \\ (Pearson/Spearman corr.)} & \makecell{MRPC \\ (acc.)} & \makecell{RTE \\ (acc.)} \\
\midrule
 & Full-precision & $84.49 / 85.20$ & $86.05 / 85.98$ & $79.11 / 79.63$ & $85.12 / 86.04$ & $86.36 / 87.02$ \\
\cmidrule{2-7}
MNLI  & Q+AdaRound & $83.69 / 84.61$ & $85.75 / 85.86$ & $72.67 / 74.11$ & $84.90 / 85.82$ & $86.33 / 86.75 $ \\
& Q+FlexRound (Ours)& $\mathbf{84.53} / \mathbf{84.98}$ & $\mathbf{85.93} / \mathbf{85.99}$ & $\mathbf{72.94} / \mathbf{74.24}$ & $\mathbf{85.56} / \mathbf{86.14}$ & $\mathbf{86.41} / \mathbf{86.89}$ \\
\midrule
& Full-precision & $88.06 / 91.08$ & $88.66 / 91.59$ & $85.20 / 88.99$ & $88.26 / 91.28$ & $88.62 / 91.50$ \\
\cmidrule{2-7}
QQP & Q+AdaRound & $87.65 / 90.58$ & $87.48 / 90.62$ & $72.97 / 79.35$ & $87.98 / 91.04$ & $88.38 / 91.27$ \\
& Q+FlexRound (Ours)& $\mathbf{87.81} / \mathbf{90.83}$ & $\mathbf{88.38} / \mathbf{91.31}$ & $\mathbf{73.75} / \mathbf{80.65}$ & $\mathbf{88.27} / \mathbf{91.18}$ & $\mathbf{88.60} / \mathbf{91.39}$ \\
\midrule
% & Full-precision & $93.00$ & $92.78$ & $89.91$ & $93.35$ & $94.50$ \\
% \cmidrule{2-7}
% SST-2 & Q+AdaRound & $\mathbf{92.66}$ & $93.00$ & $\mathbf{84.75}$ & $92.55$ & $93.81$ \\
% & Q+FlexRound (Ours)& $92.43$ & $\mathbf{93.58}$ & $83.03$ & $\mathbf{93.12}$ & $\mathbf{94.04}$ \\
% \midrule
& Full-precision & $85.05$ & $85.54$ & $80.15$ & $85.05$ & $87.99$ \\ 
\cmidrule{2-7}
MRPC & Q+AdaRound & $81.62$ & $82.35$ & $75.25$ & $84.80$ &  $85.78$ \\
& Q+FlexRound (Ours)& $\mathbf{84.07}$ & $\mathbf{84.31}$ & $\mathbf{75.49}$ & $\mathbf{85.05}$ & $\mathbf{86.76}$ \\
\bottomrule
\end{tabular}
% }
\end{center}
% \vskip -10pt
\vskip -0.15in
\end{table*}

\subsection{Ablation Study}\label{subsec:ablation}

\paragraph{Ablation Study 1} Although AdaRound demonstrates the state-of-the-art performance among previous PTQ approaches, it is unable to learn the quantization grid size $s_1$ jointly with the rounding process, as discussed in Section~\ref{sec:related}. To understand the significance of learning $s_1$ jointly with the rounding process, we evaluate the performance of FlexRound with a fixed $s_1$ (Ablation Study 1 in Table~\ref{tab:ablation}) on the ImageNet dataset with weights quantized to $4$-bit (activations are not quantized). As seen in Table~\ref{tab:ablation}, when $s_1$ is fixed, FlexRound performs similarly to AdaRound for all models except for ResNet-18. This indicates that regardless of the quantization method used, whether it be AdaRound or FlexRound, using a fixed $s_1$ prevents further improvements in the performance of the quantized model. However, when learning $s_1$ jointly with the rounding process, FlexRound outperforms AdaRound for every model. The ability to learn $s_1$ jointly with the rounding process is a critical aspect in closing the performance gap between a full-precision model and its quantized counterpart. FlexRound possesses this capability in contrast to AdaRound since it is based on element-wise division, as mentioned in Section~\ref{subsec:FlexRound}.

\paragraph{Ablation Study 2} To justify the inclusion of additional tensors $\vs_3$ and $\vs_4$ in FlexRound,
%in a per-tensor uniform PTQ setting as discussed in Section~\ref{subsec:FlexRound}
we conduct an ablation study in which 
% the impact of $\vs_3$ and $\vs_4$ on 
FlexRound is tested on the ImageNet dataset with weights quantized to $4$-bit while keeping activations unquantized, and the results are compared with FlexRound without the use of $\vs_3$ and $\vs_4$ (Ablation Study 2 in Table~\ref{tab:ablation}). As shown in the last two rows in Table~\ref{tab:ablation}, the presence of $\vs_3$ and $\vs_4$ increases the top-1 accuracy for all models. Interestingly, FlexRound without the use of $\vs_3$ and $\vs_4$ also outperforms both AdaQuant and AdaRound, which would support our claim that a new rounding scheme shifted from element-wise addition to element-wise division is the key to improving the quantization quality significantly. % All experiments in Table~\ref{tab:ablation} are done using the ImageNet dataset with pre-trained weights quantized into $4$-bit (activations are not quantized).

% To justify the introduction of $\vs_3$ and $\vs_4$ on FlexRound in the per-tensor uniform PTQ setting, we investigate the impact of $\vs_3$ and $\vs_4$ on the performance of FlexRound using the ImageNet dataset with pre-trained weights quantized into $2$-bit (activations are not quantized). As shown in the last two rows in Table \ref{tab:ablation}, the presence of $\vs_3$ and $\vs_4$ enhances the accuracy for all models. Interestingly, FlexRound outperforms both AdaQuant and AdaRound even without $\vs_3$ and $\vs_4$, which would support our claim that a new rounding scheme, shifted from element-wise addition to element-wise division, is the key to improving quantization quality significantly.

\begin{table*}[t]
% \vskip -5pt
\vskip -0.1in
\caption{Performance of GPT-Neo and OPT fine-tuned on WikiText2 and PTB, respectively. The perplexity (PPL) is employed as a performance metric. The lower PPL, the better. ``Q $+$ X" means the implementation of X in the QDrop's setting. Both weights and input activations of attention and feed-forward sub-layers are quantized to $8$-bit in a per-tensor asymmetric scheme.}\label{tab:clm_finetuned}
\begin{center}
\small
% \resizebox{0.9\linewidth}{!}{
\begin{tabular}{clcccccc}
\toprule
Dataset & \makecell{Method} & $\text{GPT-Neo}_{125\text{M}}$ & $\text{GPT-Neo}_{1.3\text{B}}$ & $\text{GPT-Neo}_{2.7\text{B}}$ & $\text{OPT}_{125\text{M}}$ & $\text{OPT}_{1.3\text{B}}$ & $\text{OPT}_{2.7\text{B}}$ \\
\midrule
& Full-precision & $21.96$ & $12.09$ & $10.78$ & $19.85$ & $11.52$ & $10.27$ \\
\cmidrule{2-8}
WikiText2 & Q+AdaRound & $30.52$ & $12.47$ & $14.09$ & $27.96$ & $12.66$ & $10.97$ \\
& Q+FlexRound (Ours)& $\mathbf{24.30}$ & $\mathbf{12.37}$ & $\mathbf{12.43}$ & $\mathbf{21.43}$ & $\mathbf{12.02}$ & $\mathbf{10.63}$ \\
\midrule
& Full-precision & $24.20$ & $16.09$ & $14.70$ & $16.50$ & $11.62$ & $10.80$ \\
\cmidrule{2-8}
PTB & Q+AdaRound & $31.40$ & $16.63$ & $19.80$ & $20.28$ & $13.00$ & $12.02$ \\
& Q+FlexRound (Ours)& $\mathbf{26.03}$ & $\mathbf{16.32}$ & $\mathbf{16.87}$ & $\mathbf{17.68}$ & $\mathbf{12.22}$ & $\mathbf{11.29}$ \\
\bottomrule
\end{tabular}
% }
\end{center}
% \vskip -10pt
\vskip -0.15in
\end{table*}

\subsection{ResNet and MobileNetV2 on ImageNet}\label{subsec:imagenet}

We quantize ResNet-18, ResNet-50, and MobileNetV2 in the low-bit PTQ reconstruction with $1024$ randomly sampled images. Linear symmetric per-tensor quantization format is assumed for quantizing weights and/or activations, whereas in contrast, \citet{li2021brecq} and \citet{wei2022qdrop} adopt linear asymmetric per-channel quantization format, which causes discrepancies between the results obtained in our own implementation of BRECQ and QDrop and those reported in \citet{li2021brecq} and \citet{wei2022qdrop}. For FlexRound, the output of each layer or block is reconstructed during $5k$ iterations while all learnable parameters (i.e., $s_1$, $\mS_2$, $\vs_3$, and $\vs_4$) are updated by using one learning rate (e.g., $4$e-$4$ for the ResNet models quantized by $3$-bit or $4$-bit, or $1$e-$3$ for the ResNet models quantized by $2$-bit and MobileNetV2). The first and last layers are quantized to $8$-bit and the batch normalization layer is folded into convolution, as in \citet{li2021brecq}. Our experiments are performed based on full-precision pre-trained models provided in the BRECQ github repository\footnote{\url{https://github.com/yhhhli/BRECQ}}, unless otherwise noted.
% the official PyTorch \citep{paszke2019pytorch} repository\footnote{\url{https://pytorch.org/vision/stable/models.html}}, 
The experiments based on full-precision pre-trained models available from the official PyTorch repository are given in Appendix~\ref{appendix:pytorch}. We report the median over five random trials. 

Assuming the quantization of weights only, we compare FlexRound with AdaRound and AdaQuant, which both utilize the principle of element-wise addition. Table~\ref{tab:imagenet_w_brecq} shows that FlexRound consistently outperforms those two addition-based rounding policies. Note that the performance of AdaQuant is inferior to that of AdaRound in Table~\ref{tab:imagenet_w_brecq}. Correspondingly, FlexRound would be compared to AdaRound only to save space hereafter. Table~\ref{tab:imagenet_wa_brecq} provides model accuracy when AdaRound and FlexRound (quantizing both weights and activations) are associated with the settings of BRECQ or QDrop.
%Table \ref{tab:imagenet_wa} also shows that FlexRound performs better than existing methods in most cases.
It is worth noting that in Table~\ref{tab:imagenet_wa_brecq}, FlexRound is particularly effective for MobileNetV2 (which includes weights of large magnitude) for the reasons explained in Section~\ref{subsec:FlexRound}.
%Particularly, as explained in Section \ref{subsec:analysis}, FlexRound outcompetes prior methods by a considerable margin in the case of MobileNetV2 due to the presence of weights of large magnitude. 
It is also interesting to see that even when both weights and activations of the ResNet models are quantized to $4$-bit under a per-tensor uniform PTQ setting, the performance degradation (compared to a full-precision pre-trained model) is negligible (less than $2\%$) in Table~\ref{tab:imagenet_wa_brecq}.
% One might wonder why the performance of AdaQuant is too poor in $2$-bit and in the case of MobileNetV2. It may result from the per-tensor PTQ setting.

\subsection{Language Models}

\begin{table}[t]
% \vskip -5pt
\vskip -0.1in
\caption{Performance of GPT-2 medium (M) and large (L) fine-tuned on WebNLG via LoRA. ``Unseen'', ``Seen'', and ``All'' represent the BLEU score for unseen, seen, and all categories in the test set of WebNLG. The higher the BLEU score, the better. ``Q $+$ X" indicates the implementation of X in the QDrop's setting. Both weights and input activations of attention and feed-forward sub-layers are quantized to $8$-bit in a per-tensor asymmetric scheme.}
\vskip -0.1in
% \vskip -10pt
\label{tab:lora}
\begin{center}
\small
\resizebox{\linewidth}{!}{
\begin{tabular}{clccc}
\toprule
Model & \makecell{Method} & Unseen & Seen & All \\
\midrule
& Full-precision (LoRA) & $47.16$ & $62.31$ & $55.43$ \\
\cmidrule{2-5}
GPT-2 M & Q+AdaRound & $45.70$ & $60.92$ & $54.05$ \\
& Q+FlexRound (Ours)& $\mathbf{46.85}$ & $\mathbf{61.83}$ & $\mathbf{55.06}$ \\
\midrule
& Full-precision (LoRA) & $48.06$ & $64.39$ & $56.97$ \\
\cmidrule{2-5}
GPT-2 L & Q+AdaRound & $48.09$ & $63.98$ & $56.75$  \\
& Q+FlexRound (Ours)& $\mathbf{48.42}$ & $\mathbf{64.47}$ & $\mathbf{57.16}$ \\
\bottomrule
\end{tabular}
}
\end{center}
% \vskip -10pt
\vskip -0.15in
\end{table}

% \begin{table}[t]
% % \vskip -12pt
% \caption{Performance of $\text{GPT-Neo}_{125\text{M}}$, $\text{GPT-Neo}_{1.3\text{B}}$, $\text{GPT-Neo}_{2.7\text{B}}$, $\text{OPT}_{125\text{M}}$, $\text{OPT}_{1.3\text{B}}$ and $\text{OPT}_{2.7\text{B}}$ on the WikiText2 and PTB datasets. The perplexity (PPL) is employed as a performance metric. The lower PPL, the better. ``Q $+$ X" means the implementation of X in the QDrop's setting.}
% \label{tab:clm}
% \begin{center}
% % \footnotesize
% \resizebox{\linewidth}{!}{
% \begin{tabular}{clcccccc}
% \Xhline{1pt}
% Dataset & \makecell{Method} & $\text{GPT-Neo}_{125\text{M}}$ & $\text{GPT-Neo}_{1.3\text{B}}$ & $\text{GPT-Neo}_{2.7\text{B}}$ & $\text{OPT}_{125\text{M}}$ & $\text{OPT}_{1.3\text{B}}$ & $\text{OPT}_{2.7\text{B}}$ \\
% \midrule
% & Full-precision & $31.54$ & $15.40$ & $13.35$ & $56.08$ & $29.76$ & $26.13$ \\
% \cline{2-8}
% WikiText2 & Q+AdaRound & $35.60$ & $15.75$ & $13.95$ & $226.48$ & $40.40$ & $47.48$ \\
% & Q+FlexRound (Ours)& $\mathbf{33.44}$ & $\mathbf{15.68}$ & $\mathbf{13.80}$ & $\mathbf{66.07}$ & $\mathbf{40.01}$ & $\mathbf{40.38}$ \\
% \hline
% & Full-precision & $64.63$ & $31.51$ & $27.22$ & $129.90$ & $76.06$ & $68.81$ \\
% \cline{2-8}
% PTB & Q+AdaRound & $70.16$ & $31.97$ & $28.24$ & $220.01$ & $103.15$ & $120.37$ \\
% & Q+FlexRound (Ours)& $\mathbf{66.62}$ & $\mathbf{31.74}$ & $\mathbf{27.68}$ & $\mathbf{145.45}$ & $\mathbf{101.81}$ & $\mathbf{106.88}$ \\
% \Xhline{1pt}
% \end{tabular}
% }
% \end{center}
% \end{table}

\begin{table*}[t]
\vskip -0.1in
\caption{Zero-shot performance of LLaMA-$33$B on $6$ common sense reasoning tasks (BoolQ, PIQA, HellaSwag, WinoGrande, ARC easy and challenge, and OBQA) and the causal language modeling task on WikiText2. The accuracy ($\%$) and the perplexity (PPL) are reported for common sense reasoning tasks and the causal language modeling task, respectively. The lower PPL, the better. ``Q $+$ X" expresses the implementation of X in the QDrop's setting. The weights of attention and feed-forward sub-layers are quantized to $8$-bit in a per-channel asymmetric format, whereas the input activations of those sub-layers are quantized to $8$-bit in a per-tensor asymmetric scheme.}
\vskip -0.1in
\label{tab:plm_llama_33b}
\begin{center}
\small
\resizebox{\linewidth}{!}{
\begin{tabular}{clcccccccc}
\toprule
Model & \makecell{Method} & BoolQ & PIQA & HellaSwag &  WinoGrande & ARC-e & ARC-c & OBQA & WikiText2\\
\midrule
% & Half-precision & $73.15$ & $77.31$ & $72.96$ & $67.09$ & $52.48$ & $41.38$ & $42.40$ & $8.90$ \\
% \cmidrule{2-10}
% LLaMA-$7$B  & Q+AdaRound & $70.12$ & $75.08$ & $69.89$ & $65.82$ & $51.47$ & $39.42$ & $39.00$ & $10.38$ \\
% & Q+FlexRound (Ours)& $\mathbf{73.76}$ & $\mathbf{76.66}$ & $\mathbf{71.75}$ & $\mathbf{67.01}$ & $\mathbf{52.31}$ & $\mathbf{40.02}$ & $\mathbf{42.20}$ & $\mathbf{9.25}$ \\
% \midrule
% & Half-precision & $68.53$ & $79.11$ & $76.23$ & $70.01$ & $59.89$ & $44.54$ & $42.20$ & $7.73$ \\
% \cmidrule{2-10}
% LLaMA-$13$B  & Q+AdaRound & $66.09$ & $76.44$ & $72.06$ & $66.30$ & $57.32$ & $43.00$ & $39.60$ & $9.07$ \\
% & Q+FlexRound (Ours)& $\mathbf{68.59}$ & $\mathbf{78.67}$ & $\mathbf{75.21}$ & $\mathbf{70.64}$ & $\mathbf{58.88}$ & $\mathbf{43.60}$ & $\mathbf{41.20}$ & $\mathbf{8.01}$ \\
% \midrule
& Half-precision & $68.38$ & $80.09$ & $79.21$ & $72.93$ & $58.92$ & $45.48$ & $42.00$ & $6.35$ \\
\cmidrule{2-10}
LLaMA-$33$B  & Q+AdaRound & $64.86$ & $74.65$ & $68.64$ & $57.93$ & $49.28$ & $36.95$ & $41.00$ & $10.39$ \\
& Q+FlexRound (Ours)& $\mathbf{69.08}$ & $\mathbf{79.16}$ & $\mathbf{77.43}$ & $\mathbf{72.53}$ & $\mathbf{56.61}$ & $\mathbf{44.97}$ & $\mathbf{44.00}$ & $\mathbf{6.82}$ \\
% \midrule
% & Half-precision & $85.96$ & $82.48$ & $82.20$ & $80.03$ & $74.87$ & $56.23$ & $47.00$ & \\
% \cmidrule{2-10}
% Five-shot  & Q+AdaRound & $85.96$ & $82.48$ & $82.20$ & $80.03$ & $74.87$ & $56.23$ & $47.00$ & \\
% & Q+FlexRound (Ours)& $\mathbf{85.32}$ & $\mathbf{80.90}$ & $\mathbf{80.52}$ & $\mathbf{78.37}$ & $\mathbf{71.72}$ & $\mathbf{53.16}$ & $\mathbf{46.80}$ & \\
\bottomrule
\end{tabular}
}
\end{center}
\vskip -0.15in
\end{table*}

All language models in this paper are based on the structure of Transformer \citep{vaswani2017attention}. To reduce the precision of such models to $8$-bit, unless otherwise stated, we employ a linear asymmetric per-tensor quantization scheme for both weights and activations. The reconstruction step for PTQ is applied to each Transformer layer, including both attention and feed-forward sub-layers.
% The PTQ scheme was formatted as per-tensor quantization with asymmetric weight and activation. Weight and activation were quantized to 8-bit respectively including the weight of the embedding layer except the last layer with randomly initialized weights was remained as full-precision. 
All weights in attention and feed-forward sub-layers are quantized to $8$-bit.
Activations are quantized to $8$-bit on-the-fly before each linear layer, while the inputs of the softmax and normalization layers remain at full-precision as suggested in \citet{zafrir2019q8bert} and \citet{zhang2020ternarybert}. We utilize pre-trained language models (PLMs) and datasets from the HuggingFace \citep{wolf-etal-2020-transformers} repository, with the exception of GPT-2 and LLaMA experiments. % models and the WebNLG dataset. % \footnote{\url{https://github.com/huggingface/transformers}}
% The experiments for BERT fine-tuned on SQuADv1 \citep{2016arXiv160605250R} are presented in Appendix~\ref{appendix:squad}.
The experiments for the question-answering task with a fine-tuned BERT on the SQuADv1 \citep{2016arXiv160605250R} dataset is presented in Appendix~\ref{appendix:squad}.

\paragraph{BERT and GPT-Neo on GLUE} \label{subsec:nlu}
We evaluate the natural language understanding (NLU) performance of FlexRound using a variety of models including $\text{BERT}_{\text{Base}}$, $\text{BERT}_{\text{Large}}$, $\text{GPT-Neo}_{125\text{M}}$, $\text{GPT-Neo}_{1.3\text{B}}$, and $\text{GPT-Neo}_{2.7\text{B}}$ fine-tuned on the GLUE benchmark. 
%Both $\text{BERT}_{\text{Base}}$ and $\text{BERT}_{\text{Large}}$ are uncased models.
We only report the experimental results on the MNLI, QQP, % SST-2, 
and MRPC datasets due to space limit. All experimental results are presented in Appendix~\ref{appendix:nlu}.
The learning rate applied to all learnable parameters ($s_1$, $\mS_2$, and $\vs_3$) is selected to be $2$e-$4$ for BERT and to be $3$e-$4$ for GPT-Neo regardless of the task to demonstrate that `Q + FlexRound' can broadly surpass `Q + AdaRound' without the need of significant efforts to select the optimal learning rate for each task. 
Reconstruction process is performed by using $1024$ random samples for $20K$ iterations. 
The last, randomly initialized layer remains in full-precision. Further experimental details are deferred to Appendix~\ref{appendix:nlu}. % for a training recipe of finetuned model and settings. 
% The dropping probability of activation for output reconstruction was set to $0.5$. 
% \textcolor{blue}{Note that the SQuAD \citep{rajpurkar2016squad} results are attached to Appendix \ref{appendix:squad}. }  
In Table~\ref{tab:glue}, we report the performance of `Q + AdaRound' and `Q + FlexRound' that are potentially promising as shown in Table~\ref{tab:imagenet_wa_brecq}.
%, we henceforth report the performance of QDrop with AdaRound and that of QDrop with FlexRound. 
We can notice that `Q + FlexRound' yields better NLU scores than `Q + AdaRound' for all % most 
models and NLU tasks.
In particular, for the MNLI and QQP datasets, `Q + FlexRound' can achieve comparable or even superior performance to a full-precision model in a per-tensor uniform PTQ setting with the exception of $\text{GPT-Neo}_{125\text{M}}$. 
% As mentioned in Section \ref{sec:related}, it is the first time to apply per-tensor uniform PTQ to language models. 

\paragraph{GPT-Neo and OPT on WikiText2 and PTB}\label{subsec:nlg}
%Last but not least, to verify whether per-tensor uniform PTQ works well for natural language generation tasks or not, 
We test the natural language generation (NLG) performance of FlexRound using fine-tuned PLMs including $\text{GPT-Neo}_{125\text{M}}$, $\text{GPT-Neo}_{1.3\text{B}}$, $\text{GPT-Neo}_{2.7\text{B}}$, $\text{OPT}_{125\text{M}}$, $\text{OPT}_{1.3\text{B}}$, and $\text{OPT}_{2.7\text{B}}$ on the WikiText2 dataset and PTB dataset. Fine-tuned PLMs (for NLG) are quantized by AdaRound and FlexRound in a per-tensor quantization manner with $128$ random samples drawn from downstream task training data.
% We test the natural language generation (NLG) performance of FlexRound on the WikiText2 and PTB datasets. Fine-tuned PLMs (for NLG) are quantized by FlexRound (in a per-tensor quantization manner) while a small amount of data of downstream tasks are used for reconstruction and evaluation.
%\textcolor{blue}{Inspired by the zero-shot evaluation of NLG tasks, we quantize a PLM itself using a small subsest of downstream task training data and evaluate its quantized counterpart on downstream task test data, since the circumstance when a downstream task training dataset is too small to fine-tune a PLM and there is no access to pre-training data could be quite frequent in real world. In such a situation, there is no choice but to employ PTQ to quantize a PLM.}
% Specifically, PLMs include $\text{GPT-Neo}_{125\text{M}}$, $\text{GPT-Neo}_{1.3\text{B}}$, $\text{GPT-Neo}_{2.7\text{B}}$, $\text{OPT}_{125\text{M}}$, $\text{OPT}_{1.3\text{B}}$ and $\text{OPT}_{2.7\text{B}}$, while $128$ downstream task training data samples are chosen at random for reconstruction. 
More details on the experimental setup are provided in Appendix~\ref{appendix:nlg_finetune}.
%including learning rate, batch size and iteration for reconstruction.
As presented in Table~\ref{tab:clm_finetuned}, it is clear that `Q + FlexRound' is superior to `Q + AdaRound' for all models and datasets, which means that FlexRound is also effective for NLG as well as image classification and NLU. Notice that even for the OPT models, the performance of `Q + FlexRound' is close to that of a full-precision model.

\paragraph{GPT-2 on WebNLG} To this point, we have applied full fine-tuning for downstream tasks to BERT, GPT-Neo, and OPT. For language models, however, there are various fine-tuning techniques \citep{houlsby2019parameter, liu2022p, hu2022lora} that can perform better with fewer trainable parameters than full fine-tuning. 
%\textcolor{red}{In order to identify whether FlexRound is well compatible with parameter-efficient adaptation or not, we quantize GPT-2 medium and large available from the LoRA \citep{hu2022lora} repository\footnote{\url{https://github.com/microsoft/LoRA}} using AdaRound and FlexRound since LoRA is one of the state-of-the-art parameter-efficient adaptation.} 
To evaluate the compatibility of FlexRound with other fine-tuning methods, we perform experiments on quantizing GPT-2 merged with LoRA \citep{hu2022lora}, one of the state-of-the-art fine-tuning methods. We choose $128$ examples from the training set of WebNLG at random for reconstruction. More experimental details are given in Appendix~\ref{appendix:webnlg}. Table~\ref{tab:lora} shows that `Q + FlexRound' excels `Q + AdaRound', and performs similarly or even better than the full-precision model with LoRA. Hence, FlexRound is also compatible with other state-of-the-art fine-tuning techniques in addition to full fine-tuning.

\paragraph{LLaMA on Common Sense Reasoning and WikiText2} 
Finally, we evaluate the zero-shot performance of LLaMA-$33$B on six common sense reasoning benchmarks and one casual language modeling task on WikiText2. It is intended to justify that LLMs can be efficiently quantized with only negligible accuracy degradation compared to half-precision baselines by block-by-block reconstructing output, without assuming that the outliers in activations would emerge in a certain pattern. In Table~\ref{tab:plm_llama_33b}, for reconstruction, $512$ samples are randomly selected from the training dataset of C4 \citep{raffel2020c4}. We use linear asymmetric per-channel quantization for weights but linear asymmetric per-tensor quantization for activations. The zero-shot and five-shot performances of LLaMA-$7$B, LLaMA-$13$B, and LLaMA-$33$B as well as those experimental details are given in Appendix~\ref{appendix:llama}. Table~\ref{tab:plm_llama_33b} shows that `Q + FlexRound' can maintain the accuracy of the half-precision baseline, surpassing `Q + AdaRound'. Without any assumption about the activation outliers in LLMs, FlexRound can quantize LLMs while preserving the performance of half-precision baselines.

\section{Conclusion}\label{sec:conclusion}
We propose a new rounding scheme, \emph{FlexRound}, for post-training weight quantization under the principle of element-wise division, to enable jointly learning both a common quantization grid size and an individual scale for each pre-trained weight. We validate that FlexRound can flexibly quantize pre-trained weights by updating their corresponding scales depending on their own magnitudes.
% as a metric to measure importance. 
Hence, FlexRound can be applied to various models including even large language models with negligible accuracy degradation.
% achieve comparable performance to a full-precision model even in a per-tensor uniform PTQ setting. 
% \textcolor{blue}{As a future work, we plan to quantize large language models in a per-tensor uniform PTQ setting.}
% However, we cannot yet figure out the reason why FlexRound cannot perform as well as a full-precision pre-trained model. Besides, we cannot test FlexRound on large language models beyond $6.7$B parameters due to limited resources. As a future work, we plan to study how to quantize the OPT models well and quantize such large language models in the per-tensor uniform PTQ setting.
% limitations: relation with block unit size and error accumulation, large model and resource limitation, 

\nocite{langley00}

\newpage

\bibliography{references}

\begin{thebibliography}{54}
\providecommand{\natexlab}[1]{#1}
\providecommand{\url}[1]{\texttt{#1}}
\expandafter\ifx\csname urlstyle\endcsname\relax
  \providecommand{\doi}[1]{doi: #1}\else
  \providecommand{\doi}{doi: \begingroup \urlstyle{rm}\Url}\fi

\bibitem[Bai et~al.(2021)Bai, Hou, Shang, Jiang, King, and Lyu]{bai2021towards}
Bai, H., Hou, L., Shang, L., Jiang, X., King, I., and Lyu, M.~R.
\newblock Towards efficient post-training quantization of pre-trained language
  models.
\newblock \emph{arXiv preprint arXiv:2109.15082}, 2021.

\bibitem[Bengio et~al.(2013)Bengio, Leonard, and Courville]{Bengio2013ste}
Bengio, Y., Leonard, N., and Courville, A.
\newblock Estimating or propagating gradients through stochastic neurons for
  conditional computation.
\newblock \emph{arXiv preprint arXiv:1308.3432}, 2013.

\bibitem[Bisk et~al.(2020)Bisk, Zellers, Gao, Choi, et~al.]{bisk2020piqa}
Bisk, Y., Zellers, R., Gao, J., Choi, Y., et~al.
\newblock Piqa: Reasoning about physical commonsense in natural language.
\newblock In \emph{Proceedings of the AAAI conference on artificial
  intelligence}, volume~34, pp.\  7432--7439, 2020.

\bibitem[Black et~al.(2021)Black, Gao, Wang, Leahy, and Biderman]{gpt-neo}
Black, S., Gao, L., Wang, P., Leahy, C., and Biderman, S.
\newblock {GPT-Neo: Large Scale Autoregressive Language Modeling with
  Mesh-Tensorflow}, March 2021.
\newblock URL \url{https://doi.org/10.5281/zenodo.5297715}.
\newblock {If you use this software, please cite it using these metadata.}

\bibitem[Bondarenko et~al.(2021)Bondarenko, Nagel, and
  Blankevoort]{bondarenko2021understanding}
Bondarenko, Y., Nagel, M., and Blankevoort, T.
\newblock Understanding and overcoming the challenges of efficient transformer
  quantization.
\newblock In \emph{Proceedings of the 2021 Conference on Empirical Methods in
  Natural Language Processing}, pp.\  7947--7969. Association for Computational
  Linguistics, November 2021.
\newblock \doi{10.18653/v1/2021.emnlp-main.627}.
\newblock URL \url{https://aclanthology.org/2021.emnlp-main.627}.

\bibitem[Clark et~al.(2019)Clark, Lee, Chang, Kwiatkowski, Collins, and
  Toutanova]{clark2019boolq}
Clark, C., Lee, K., Chang, M.-W., Kwiatkowski, T., Collins, M., and Toutanova,
  K.
\newblock {B}ool{Q}: Exploring the surprising difficulty of natural yes/no
  questions.
\newblock In \emph{Proceedings of the 2019 Conference of the North {A}merican
  Chapter of the Association for Computational Linguistics: Human Language
  Technologies, Volume 1 (Long and Short Papers)}, pp.\  2924--2936,
  Minneapolis, Minnesota, June 2019. Association for Computational Linguistics.
\newblock \doi{10.18653/v1/N19-1300}.
\newblock URL \url{https://aclanthology.org/N19-1300}.

\bibitem[Clark et~al.(2018)Clark, Cowhey, Etzioni, Khot, Sabharwal, Schoenick,
  and Tafjord]{clark2018arc}
Clark, P., Cowhey, I., Etzioni, O., Khot, T., Sabharwal, A., Schoenick, C., and
  Tafjord, O.
\newblock Think you have solved question answering? try arc, the ai2 reasoning
  challenge.
\newblock \emph{arXiv preprint arXiv:1803.05457}, 2018.

\bibitem[Courbariaux et~al.(2016)Courbariaux, Hubara, Soudry, El-Yaniv, and
  Bengio]{courbariaux2016binarized}
Courbariaux, M., Hubara, I., Soudry, D., El-Yaniv, R., and Bengio, Y.
\newblock Binarized neural networks: Training deep neural networks with weights
  and activations constrained to +1 or-1.
\newblock \emph{arXiv preprint arXiv:1602.02830}, 2016.

\bibitem[Dettmers et~al.(2022)Dettmers, Lewis, Belkada, and
  Zettlemoyer]{dettmers2022llm}
Dettmers, T., Lewis, M., Belkada, Y., and Zettlemoyer, L.
\newblock Llm. int8 (): 8-bit matrix multiplication for transformers at scale.
\newblock \emph{arXiv preprint arXiv:2208.07339}, 2022.

\bibitem[Devlin et~al.(2018)Devlin, Chang, Lee, and Toutanova]{devlin2018bert}
Devlin, J., Chang, M.-W., Lee, K., and Toutanova, K.
\newblock Bert: Pre-training of deep bidirectional transformers for language
  understanding.
\newblock \emph{arXiv preprint arXiv:1810.04805}, 2018.

\bibitem[Esser et~al.(2020)Esser, McKinstry, Bablani, Appuswamy, and
  Modha]{esser2020learned}
Esser, S.~K., McKinstry, J.~L., Bablani, D., Appuswamy, R., and Modha, D.~S.
\newblock Learned step size quantization.
\newblock In \emph{International Conference on Learning Representations}, 2020.
\newblock URL \url{https://openreview.net/forum?id=rkgO66VKDS}.

\bibitem[Gao et~al.(2021)Gao, Tow, Biderman, Black, DiPofi, Foster, Golding,
  Hsu, McDonell, Muennighoff, Phang, Reynolds, Tang, Thite, Wang, Wang, and
  Zou]{eval-harness}
Gao, L., Tow, J., Biderman, S., Black, S., DiPofi, A., Foster, C., Golding, L.,
  Hsu, J., McDonell, K., Muennighoff, N., Phang, J., Reynolds, L., Tang, E.,
  Thite, A., Wang, B., Wang, K., and Zou, A.
\newblock A framework for few-shot language model evaluation, September 2021.
\newblock URL \url{https://doi.org/10.5281/zenodo.5371628}.

\bibitem[Gardent et~al.(2017)Gardent, Shimorina, Narayan, and
  Perez-Beltrachini]{gardent2017webnlg}
Gardent, C., Shimorina, A., Narayan, S., and Perez-Beltrachini, L.
\newblock The {W}eb{NLG} challenge: Generating text from {RDF} data.
\newblock In \emph{Proceedings of the 10th International Conference on Natural
  Language Generation}, pp.\  124--133, Santiago de Compostela, Spain,
  September 2017. Association for Computational Linguistics.
\newblock \doi{10.18653/v1/W17-3518}.
\newblock URL \url{https://aclanthology.org/W17-3518}.

\bibitem[Han et~al.(2015)Han, Pool, Tran, and Dally]{han2015learning}
Han, S., Pool, J., Tran, J., and Dally, W.
\newblock Learning both weights and connections for efficient neural network.
\newblock \emph{Advances in neural information processing systems}, 28, 2015.

\bibitem[Han et~al.(2016)Han, Mao, and Dally]{han2016deep}
Han, S., Mao, H., and Dally, W.~J.
\newblock Deep compression: Compressing deep neural networks with pruning,
  trained quantization and huffman coding.
\newblock In \emph{International Conference on Learning Representations}, 2016.
\newblock URL \url{https://arxiv.org/pdf/1510.00149.pdf}.

\bibitem[He et~al.(2016)He, Zhang, Ren, and Sun]{he2016deep}
He, K., Zhang, X., Ren, S., and Sun, J.
\newblock Deep residual learning for image recognition.
\newblock In \emph{Proceedings of the IEEE conference on computer vision and
  pattern recognition}, pp.\  770--778, 2016.

\bibitem[Houlsby et~al.(2019)Houlsby, Giurgiu, Jastrzebski, Morrone,
  De~Laroussilhe, Gesmundo, Attariyan, and Gelly]{houlsby2019parameter}
Houlsby, N., Giurgiu, A., Jastrzebski, S., Morrone, B., De~Laroussilhe, Q.,
  Gesmundo, A., Attariyan, M., and Gelly, S.
\newblock Parameter-efficient transfer learning for nlp.
\newblock In \emph{International Conference on Machine Learning}, pp.\
  2790--2799. PMLR, 2019.

\bibitem[Hu et~al.(2022)Hu, yelong shen, Wallis, Allen-Zhu, Li, Wang, Wang, and
  Chen]{hu2022lora}
Hu, E.~J., yelong shen, Wallis, P., Allen-Zhu, Z., Li, Y., Wang, S., Wang, L.,
  and Chen, W.
\newblock Lo{RA}: Low-rank adaptation of large language models.
\newblock In \emph{International Conference on Learning Representations}, 2022.
\newblock URL \url{https://openreview.net/forum?id=nZeVKeeFYf9}.

\bibitem[Hubara et~al.(2021)Hubara, Nahshan, Hanani, Banner, and
  Soudry]{hubara2021adaquant}
Hubara, I., Nahshan, Y., Hanani, Y., Banner, R., and Soudry, D.
\newblock Accurate post training quantization with small calibration sets.
\newblock In \emph{Proceedings of the 38th International Conference on Machine
  Learning}, volume 139 of \emph{Proceedings of Machine Learning Research},
  pp.\  4466--4475. PMLR, 2021.
\newblock URL \url{https://proceedings.mlr.press/v139/hubara21a.html}.

\bibitem[Jain et~al.(2019)Jain, Gural, Wu, and Dick]{jain2019tqt}
Jain, S.~R., Gural, A., Wu, M., and Dick, C.~H.
\newblock Trained quantization thresholds for accurate and efficient
  fixed-point inference of deep neural networks.
\newblock \emph{arXiv preprint arXiv:1903.08066}, 2019.

\bibitem[Jung et~al.(2019)Jung, Son, Lee, Son, Han, Kwak, Ju~Hwang, and
  Choi]{jung2019learning}
Jung, S., Son, C., Lee, S., Son, J., Han, J.-J., Kwak, Y., Ju~Hwang, S., and
  Choi, C.
\newblock Learning to quantize deep networks by optimizing quantization
  intervals with task loss.
\newblock In \emph{The IEEE Conference on Computer Vision and Pattern
  Recognition (CVPR)}, pp.\  4350--4359, 2019.

\bibitem[Kim et~al.(2021)Kim, Park, and Yi]{kim2021performance}
Kim, S., Park, G., and Yi, Y.
\newblock Performance evaluation of int8 quantized inference on mobile gpus.
\newblock \emph{IEEE Access}, 9:\penalty0 164245--164255, 2021.

\bibitem[Lee et~al.(2021)Lee, Yun, Hwang, and Yang]{lee2021cluster}
Lee, J.~H., Yun, J., Hwang, S.~J., and Yang, E.
\newblock Cluster-promoting quantization with bit-drop for minimizing network
  quantization loss.
\newblock In \emph{2021 IEEE/CVF International Conference on Computer Vision
  (ICCV)}, pp.\  5350--5359. IEEE Computer Society, 2021.
\newblock URL
  \url{https://doi.ieeecomputersociety.org/10.1109/ICCV48922.2021.00532}.

\bibitem[Li et~al.(2021)Li, Gong, Tan, Yang, Hu, Zhang, Yu, Wang, and
  Gu]{li2021brecq}
Li, Y., Gong, R., Tan, X., Yang, Y., Hu, P., Zhang, Q., Yu, F., Wang, W., and
  Gu, S.
\newblock {BRECQ}: Pushing the limit of post-training quantization by block
  reconstruction.
\newblock In \emph{International Conference on Learning Representations}, 2021.
\newblock URL \url{https://openreview.net/forum?id=POWv6hDd9XH}.

\bibitem[Liu et~al.(2022)Liu, Ji, Fu, Tam, Du, Yang, and Tang]{liu2022p}
Liu, X., Ji, K., Fu, Y., Tam, W., Du, Z., Yang, Z., and Tang, J.
\newblock P-tuning: Prompt tuning can be comparable to fine-tuning across
  scales and tasks.
\newblock In \emph{Proceedings of the 60th Annual Meeting of the Association
  for Computational Linguistics (Volume 2: Short Papers)}, pp.\  61--68, 2022.

\bibitem[Lou et~al.(2020)Lou, Guo, Kim, Liu, and Jiang.]{lou2020autoq}
Lou, Q., Guo, F., Kim, M., Liu, L., and Jiang., L.
\newblock Autoq: Automated kernel-wise neural network quantization.
\newblock In \emph{International Conference on Learning Representations}, 2020.
\newblock URL \url{https://openreview.net/forum?id=rygfnn4twS}.

\bibitem[Marcus et~al.(1993)Marcus, Santorini, and
  Marcinkiewicz]{marcus-etal-1993-building}
Marcus, M.~P., Santorini, B., and Marcinkiewicz, M.~A.
\newblock Building a large annotated corpus of {E}nglish: The {P}enn
  {T}reebank.
\newblock \emph{Computational Linguistics}, 19\penalty0 (2):\penalty0 313--330,
  1993.
\newblock URL \url{https://www.aclweb.org/anthology/J93-2004}.

\bibitem[Merity et~al.(2016)Merity, Xiong, Bradbury, and
  Socher]{merity2016pointer}
Merity, S., Xiong, C., Bradbury, J., and Socher, R.
\newblock Pointer sentinel mixture models, 2016.

\bibitem[Mihaylov et~al.(2018)Mihaylov, Clark, Khot, and
  Sabharwal]{mihaylov2018obqa}
Mihaylov, T., Clark, P., Khot, T., and Sabharwal, A.
\newblock Can a suit of armor conduct electricity? a new dataset for open book
  question answering.
\newblock \emph{arXiv preprint arXiv:1809.02789}, 2018.

\bibitem[Nagel et~al.(2019)Nagel, Baalen, Blankevoort, and
  Welling]{nagel2019data}
Nagel, M., Baalen, M.~v., Blankevoort, T., and Welling, M.
\newblock Data-free quantization through weight equalization and bias
  correction.
\newblock In \emph{Proceedings of the IEEE/CVF International Conference on
  Computer Vision}, pp.\  1325--1334, 2019.

\bibitem[Nagel et~al.(2020)Nagel, Amjad, Van~Baalen, Louizos, and
  Blankevoort]{nagel2020adaround}
Nagel, M., Amjad, R.~A., Van~Baalen, M., Louizos, C., and Blankevoort, T.
\newblock Up or down? {A}daptive rounding for post-training quantization.
\newblock In \emph{Proceedings of the 37th International Conference on Machine
  Learning}, volume 119 of \emph{Proceedings of Machine Learning Research},
  pp.\  7197--7206. PMLR, 2020.
\newblock URL \url{https://proceedings.mlr.press/v119/nagel20a.html}.

\bibitem[Nagel et~al.(2021)Nagel, Fournarakis, Amjad, Bondarenko, van Baalen,
  and Blankevoort]{nagel2021white}
Nagel, M., Fournarakis, M., Amjad, R.~A., Bondarenko, Y., van Baalen, M., and
  Blankevoort, T.
\newblock A white paper on neural network quantization.
\newblock \emph{arXiv preprint arXiv:2106.08295}, 2021.

\bibitem[Nahshan et~al.(2021)Nahshan, Chmiel, Baskin, Zheltonozhskii, Banner,
  Bronstein, and Mendelson]{nahshan2021loss}
Nahshan, Y., Chmiel, B., Baskin, C., Zheltonozhskii, E., Banner, R., Bronstein,
  A.~M., and Mendelson, A.
\newblock Loss aware post-training quantization.
\newblock \emph{Machine Learning}, 110\penalty0 (11):\penalty0 3245--3262,
  2021.

\bibitem[Radford et~al.(2019)Radford, Wu, Child, Luan, Amodei, Sutskever,
  et~al.]{radford2019gpt2}
Radford, A., Wu, J., Child, R., Luan, D., Amodei, D., Sutskever, I., et~al.
\newblock Language models are unsupervised multitask learners.
\newblock \emph{OpenAI blog}, 1\penalty0 (8):\penalty0 9, 2019.

\bibitem[Raffel et~al.(2020)Raffel, Shazeer, Roberts, Lee, Narang, Matena,
  Zhou, Li, and Liu]{raffel2020c4}
Raffel, C., Shazeer, N., Roberts, A., Lee, K., Narang, S., Matena, M., Zhou,
  Y., Li, W., and Liu, P.~J.
\newblock Exploring the limits of transfer learning with a unified text-to-text
  transformer.
\newblock \emph{The Journal of Machine Learning Research}, 21\penalty0
  (1):\penalty0 5485--5551, 2020.

\bibitem[{Rajpurkar} et~al.(2016){Rajpurkar}, {Zhang}, {Lopyrev}, and
  {Liang}]{2016arXiv160605250R}
{Rajpurkar}, P., {Zhang}, J., {Lopyrev}, K., and {Liang}, P.
\newblock {SQuAD: 100,000+ Questions for Machine Comprehension of Text}.
\newblock \emph{arXiv e-prints}, art. arXiv:1606.05250, 2016.

\bibitem[Russakovsky et~al.(2015)Russakovsky, Deng, Su, Krause, Satheesh, Ma,
  Huang, Karpathy, Khosla, Bernstein, et~al.]{russakovsky2015imagenet}
Russakovsky, O., Deng, J., Su, H., Krause, J., Satheesh, S., Ma, S., Huang, Z.,
  Karpathy, A., Khosla, A., Bernstein, M., et~al.
\newblock Imagenet large scale visual recognition challenge.
\newblock \emph{International journal of computer vision}, 115\penalty0
  (3):\penalty0 211--252, 2015.

\bibitem[Sakaguchi et~al.(2021)Sakaguchi, Bras, Bhagavatula, and
  Choi]{sakaguchi2021winogrande}
Sakaguchi, K., Bras, R.~L., Bhagavatula, C., and Choi, Y.
\newblock Winogrande: An adversarial winograd schema challenge at scale.
\newblock \emph{Communications of the ACM}, 64\penalty0 (9):\penalty0 99--106,
  2021.

\bibitem[Sandler et~al.(2018)Sandler, Howard, Zhu, Zhmoginov, and
  Chen]{sandler2018mobilenetv2}
Sandler, M., Howard, A., Zhu, M., Zhmoginov, A., and Chen, L.-C.
\newblock Mobilenetv2: Inverted residuals and linear bottlenecks.
\newblock In \emph{Proceedings of the IEEE conference on computer vision and
  pattern recognition}, pp.\  4510--4520, 2018.

\bibitem[Touvron et~al.(2023)Touvron, Lavril, Izacard, Martinet, Lachaux,
  Lacroix, Rozière, Goyal, Hambro, Azhar, Rodriguez, Joulin, Grave, and
  Lample]{touvron2023llama}
Touvron, H., Lavril, T., Izacard, G., Martinet, X., Lachaux, M.-A., Lacroix,
  T., Rozière, B., Goyal, N., Hambro, E., Azhar, F., Rodriguez, A., Joulin,
  A., Grave, E., and Lample, G.
\newblock Llama: Open and efficient foundation language models, 2023.

\bibitem[Vaswani et~al.(2017)Vaswani, Shazeer, Parmar, Uszkoreit, Jones, Gomez,
  Kaiser, and Polosukhin]{vaswani2017attention}
Vaswani, A., Shazeer, N., Parmar, N., Uszkoreit, J., Jones, L., Gomez, A.~N.,
  Kaiser, {\L}., and Polosukhin, I.
\newblock Attention is all you need.
\newblock \emph{Advances in neural information processing systems}, 30, 2017.

\bibitem[Wang et~al.(2018)Wang, Singh, Michael, Hill, Levy, and
  Bowman]{wang2018glue}
Wang, A., Singh, A., Michael, J., Hill, F., Levy, O., and Bowman, S.~R.
\newblock Glue: A multi-task benchmark and analysis platform for natural
  language understanding.
\newblock \emph{arXiv preprint arXiv:1804.07461}, 2018.

\bibitem[Wang et~al.(2020)Wang, Chen, He, and Cheng]{wang2020towards}
Wang, P., Chen, Q., He, X., and Cheng, J.
\newblock Towards accurate post-training network quantization via bit-split and
  stitching.
\newblock In \emph{International Conference on Machine Learning}, pp.\
  9847--9856. PMLR, 2020.

\bibitem[Wei et~al.(2022)Wei, Gong, Li, Liu, and Yu]{wei2022qdrop}
Wei, X., Gong, R., Li, Y., Liu, X., and Yu, F.
\newblock {QD}rop: Randomly dropping quantization for extremely low-bit
  post-training quantization.
\newblock In \emph{International Conference on Learning Representations}, 2022.
\newblock URL \url{https://openreview.net/forum?id=ySQH0oDyp7}.

\bibitem[Wolf et~al.(2020)Wolf, Debut, Sanh, Chaumond, Delangue, Moi, Cistac,
  Rault, Louf, Funtowicz, Davison, Shleifer, von Platen, Ma, Jernite, Plu, Xu,
  Le~Scao, Gugger, Drame, Lhoest, and Rush]{wolf-etal-2020-transformers}
Wolf, T., Debut, L., Sanh, V., Chaumond, J., Delangue, C., Moi, A., Cistac, P.,
  Rault, T., Louf, R., Funtowicz, M., Davison, J., Shleifer, S., von Platen,
  P., Ma, C., Jernite, Y., Plu, J., Xu, C., Le~Scao, T., Gugger, S., Drame, M.,
  Lhoest, Q., and Rush, A.
\newblock Transformers: State-of-the-art natural language processing.
\newblock In \emph{Proceedings of the 2020 Conference on Empirical Methods in
  Natural Language Processing: System Demonstrations}, pp.\  38--45, Online,
  October 2020. Association for Computational Linguistics.
\newblock \doi{10.18653/v1/2020.emnlp-demos.6}.
\newblock URL \url{https://aclanthology.org/2020.emnlp-demos.6}.

\bibitem[Wu et~al.(2020)Wu, Judd, Zhang, Isaev, and
  Micikevicius]{wu2020integer}
Wu, H., Judd, P., Zhang, X., Isaev, M., and Micikevicius, P.
\newblock Integer quantization for deep learning inference: Principles and
  empirical evaluation.
\newblock \emph{arXiv preprint arXiv:2004.09602}, 2020.

\bibitem[Xiao et~al.(2022)Xiao, Lin, Seznec, Demouth, and
  Han]{xiao2022smoothquant}
Xiao, G., Lin, J., Seznec, M., Demouth, J., and Han, S.
\newblock Smoothquant: Accurate and efficient post-training quantization for
  large language models.
\newblock \emph{arXiv preprint arXiv:2211.10438}, 2022.

\bibitem[Yao et~al.(2022)Yao, Aminabadi, Zhang, Wu, Li, and
  He]{yao2022zeroquant}
Yao, Z., Aminabadi, R.~Y., Zhang, M., Wu, X., Li, C., and He, Y.
\newblock Zeroquant: Efficient and affordable post-training quantization for
  large-scale transformers.
\newblock \emph{arXiv preprint arXiv:2206.01861}, 2022.

\bibitem[Zafrir et~al.(2019)Zafrir, Boudoukh, Izsak, and
  Wasserblat]{zafrir2019q8bert}
Zafrir, O., Boudoukh, G., Izsak, P., and Wasserblat, M.
\newblock Q8bert: Quantized 8bit bert.
\newblock In \emph{2019 Fifth Workshop on Energy Efficient Machine Learning and
  Cognitive Computing-NeurIPS Edition (EMC2-NIPS)}, pp.\  36--39. IEEE, 2019.

\bibitem[Zellers et~al.(2019)Zellers, Holtzman, Bisk, Farhadi, and
  Choi]{zellers2019hellaswag}
Zellers, R., Holtzman, A., Bisk, Y., Farhadi, A., and Choi, Y.
\newblock Hellaswag: Can a machine really finish your sentence?
\newblock \emph{arXiv preprint arXiv:1905.07830}, 2019.

\bibitem[Zhang et~al.(2022)Zhang, Roller, Goyal, Artetxe, Chen, Chen, Dewan,
  Diab, Li, Lin, et~al.]{zhang2022opt}
Zhang, S., Roller, S., Goyal, N., Artetxe, M., Chen, M., Chen, S., Dewan, C.,
  Diab, M., Li, X., Lin, X.~V., et~al.
\newblock Opt: Open pre-trained transformer language models.
\newblock \emph{arXiv preprint arXiv:2205.01068}, 2022.

\bibitem[Zhang et~al.(2020)Zhang, Hou, Yin, Shang, Chen, Jiang, and
  Liu]{zhang2020ternarybert}
Zhang, W., Hou, L., Yin, Y., Shang, L., Chen, X., Jiang, X., and Liu, Q.
\newblock Ternarybert: Distillation-aware ultra-low bit bert.
\newblock \emph{arXiv preprint arXiv:2009.12812}, 2020.

\bibitem[Zhao et~al.(2019)Zhao, Hu, Dotzel, De~Sa, and
  Zhang]{zhao2019improving}
Zhao, R., Hu, Y., Dotzel, J., De~Sa, C., and Zhang, Z.
\newblock Improving neural network quantization without retraining using
  outlier channel splitting.
\newblock In \emph{International conference on machine learning}, pp.\
  7543--7552. PMLR, 2019.

\bibitem[Zhao et~al.(2020)Zhao, Wang, Cai, Liu, and Zhang]{zhao2020linear}
Zhao, X., Wang, Y., Cai, X., Liu, C., and Zhang, L.
\newblock Linear symmetric quantization of neural networks for low-precision
  integer hardware.
\newblock In \emph{International Conference on Learning Representations}, 2020.
\newblock URL \url{https://openreview.net/forum?id=H1lBj2VFPS}.

\end{thebibliography}
\bibliographystyle{icml2023}

%%%%%%%%%%%%%%%%%%%%%%%%%%%%%%%%%%%%%%%%%%%%%%%%%%%%%%%%%%%%%%%%%%%%%%%%%%%%%%%
%%%%%%%%%%%%%%%%%%%%%%%%%%%%%%%%%%%%%%%%%%%%%%%%%%%%%%%%%%%%%%%%%%%%%%%%%%%%%%%
% APPENDIX
%%%%%%%%%%%%%%%%%%%%%%%%%%%%%%%%%%%%%%%%%%%%%%%%%%%%%%%%%%%%%%%%%%%%%%%%%%%%%%%
%%%%%%%%%%%%%%%%%%%%%%%%%%%%%%%%%%%%%%%%%%%%%%%%%%%%%%%%%%%%%%%%%%%%%%%%%%%%%%%
\newpage
\appendix
\onecolumn
\section{Comparison of Rounding Results of AdaRound, AdaQuant, and FlexRound}\label{appendix:comparison}

\begin{figure}[h]
    \centering
    \subfigure[MobileNetV2]
    {
        \includegraphics[width=0.9\linewidth]{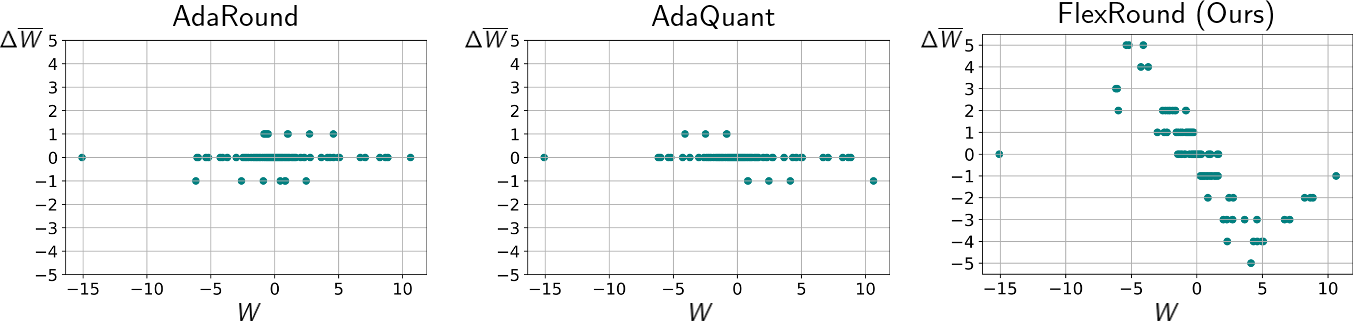}
        \label{fig:mobilenet_appendix}
    }
    \\
    \subfigure[ResNet-18]
    {
        \includegraphics[width=0.9\linewidth]{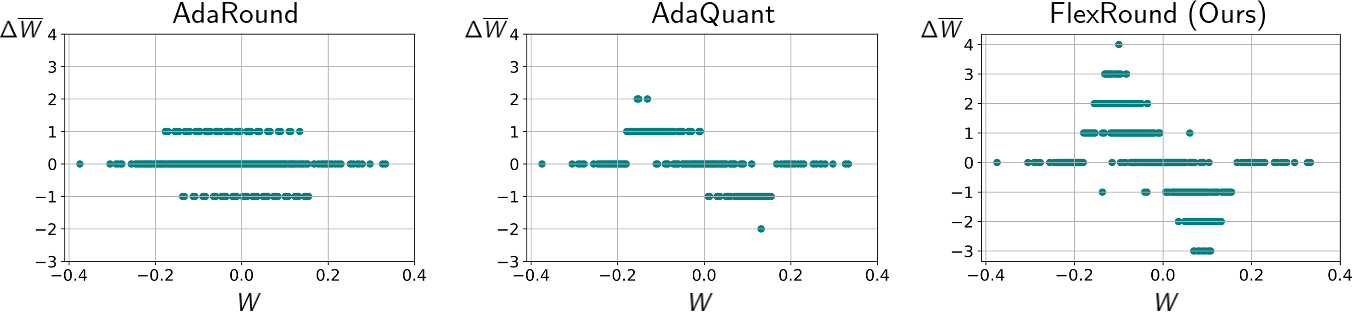}
        \label{fig:resnet_appendix}
    }
    \caption{Scatter plot of the amount of grid shifts from rounding-to-nearest grid in the first layer of the first block in MobileNetV2 and ResNet-18 when only weights are quantized to $4$-bit.} %  In relation with Figure \ref{fig:histogram}, scatter plot shows that the FlexRound can flexibly quantize the pre-trained weight well. 
    \label{fig:scatter_comparison}
\end{figure}

Figure~\ref{fig:scatter_comparison} shows the comparison of rounding results of AdaRound, AdaQuant, and FlexRound. As shown in Figure~\ref{fig:mobilenet_appendix}, FlexRound can quantize pre-trained weights more flexibly than AdaRound and AdaQuant for both ResNet-18 and MobileNetV2, thereby obtaining better performance than AdaRound and AdaQuant.  % utilize wider quantization grid than AdaRound and AdaQuant. Even in the ResNet-18 in Figure \ref{fig:resnet_appendix}, quantization grid of FlexRound is more wider than others.
% As weights of large magnitude are not quantized aggressively in the middle of Figure \ref{fig:mobilenet_appendix} compared to the right of Figure \ref{fig:mobilenet_appendix}, AdaQuant quantizes weights of large importance marginally, which seems to make it difficult for AdaQuant to quantize MobileNetV2 into $4$-bit.

\newpage

\section{Proof of Proposition~\ref{prop}}\label{appendix:proof}
Let $\mS'$ be $\mS_2 \odot \vs_3$ for a linear layer, or $\mS_2 \odot \vs_3 \odot \vs_4$ for a 2D convolution. 

For a linear layer,
\begin{align*}
    {{\partial\mathcal{L}} \over {\partial S'_{(i, j)}}} &= {{\partial \widehat{W}_{(i, j)}} \over {\partial S'_{(i, j)}}} {{\partial\mathcal{L}} \over {\partial \widehat{W}_{(i, j)}}} \\ 
    &= {{\partial \over {\partial S'_{(i, j)}}}}\Big(s_1 \Big\lfloor {W_{(i, j)} \over {s_1 S'_{(i, j)}}} \Big\rceil \Big) {{\partial\mathcal{L}} \over {\partial \widehat{W}_{(i, j)}}} \\
    &= s_1 {{\partial \over {\partial S'_{(i, j)}}}}\Big(\Big\lfloor {W_{(i, j)} \over {s_1 S'_{(i, j)}}} \Big\rceil \Big) {{\partial\mathcal{L}} \over {\partial \widehat{W}_{(i, j)}}} \\
    &= s_1 {{\partial \over {\partial S'_{(i, j)}}}} \Big( {W_{(i, j)} \over {s_1 S'_{(i, j)}}} \Big) {{\partial\mathcal{L}} \over {\partial \widehat{W}_{(i, j)}}} \quad (\because \text{Straight-Through Estimator}) \\
    &= s_1 {W_{(i, j)} \over s_1} {{\partial \over {\partial S'_{(i, j)}}}} \Big( {1 \over {S'_{(i, j)}}} \Big) {{\partial\mathcal{L}} \over {\partial \widehat{W}_{(i, j)}}} \\
    &= W_{(i, j)} \Big( - {1 \over {S'^2_{(i, j)}}} \Big) {{\partial\mathcal{L}} \over {\partial \widehat{W}_{(i, j)}}} \\ 
    &= - {W_{(i, j)} \over {S'^2_{(i, j)}}} {{\partial\mathcal{L}} \over {\partial \widehat{W}_{(i, j)}}}.
\end{align*}

For a 2D convolution, Proposition~\ref{prop} can be proved by just replacing $\widehat{W}_{(i, j)}$ and $S'_{(i, j)}$ with $\widehat{W}_{(i, j, k, l)}$ and $S'_{(i, j, k, l)}$, respectively.

\newpage

\section{ResNet-18, ResNet-50, and MobileNetV2 on ImageNet with Pre-trained Models from the Official PyTorch Repository\protect\footnote{\url{https://pytorch.org/vision/stable/models.html}}}\label{appendix:pytorch}

\begin{table}[h]
\caption{Top-1/Top-5 accuracy (\%) on ImageNet when only weights are quantized. ``B $+$ X" expresses the implementation of X in the BRECQ's setting. Here, we employ pre-trained models available from the official PyTorch repository.}
\label{tab:imagenet_w}
\begin{center}
\small
\begin{tabular}{lcccc}
\toprule
\makecell{Method} & \makecell{\# Bits (W/A)} & \makecell{ResNet-18} & \makecell{ResNet-50} & \makecell{MobileNetV2} \\
\midrule
Full-precision & $32 / 32$ & $69.76 / 89.08$ & $76.15 / 92.87$ & $71.88 / 90.29$\\
\midrule
B + AdaQuant & $4 / 32$ & $67.55 / 87.73$ & $74.09 / 91.77$ & $0.48 / 0.53$  \\
B + AdaRound & $4 / 32$ & $69.15 / 88.70$ & $75.51 / 92.73$ & $67.76 / 88.12$ \\
B + FlexRound (Ours) & $4 / 32$ & $\mathbf{69.21} / \mathbf{88.76}$ & $\mathbf{75.59} / \mathbf{92.63}$ & $\mathbf{69.56} / \mathbf{89.02}$ \\ 
\midrule
B + AdaQuant & $3 / 32$ & $60.75 / 83.41$ & $66.19 / 87.08$ & $0.10 / 0.52$  \\
B + AdaRound & $3 / 32$ & $67.98 / 88.17$ & $74.51 / 92.20$ & $60.18 / 83.52$ \\
B + FlexRound (Ours)& $3 / 32$ & $\mathbf{68.02} / \mathbf{88.03}$ & $\mathbf{74.61} / \mathbf{92.11}$ & $\mathbf{64.85} / \mathbf{86.38}$ \\
\midrule
B + AdaQuant & $2 / 32$ & $1.13 / 4.10$ & $0.12 / 0.60$ & $0.10 / 0.50$  \\
B + AdaRound & $2 / 32$ & $63.01 / 85.20$ & $68.31 / 88.98$ & $33.10 / 60.58$ \\
B + FlexRound (Ours)& $2 / 32$ & $\mathbf{63.73} / \mathbf{85.41}$ & $\mathbf{70.57} / \mathbf{90.07}$ & $\mathbf{38.09} / \mathbf{64.90}$ \\
\bottomrule
\end{tabular}
\end{center}
\end{table}

\begin{table}[h]
% \vskip -5pt
\caption{Top-1/Top-5 accuracy (\%) on ImageNet when both weights and activations are quantized. ``B $+$ X" and ``Q $+$ Y" represent the implementation of X in the BRECQ's setting and that of Y in the QDrop's setting, respectively. Here, we employ pre-trained models available from the official PyTorch repository.}
\label{tab:imagenet_wa}
\begin{center}
\small
\begin{tabular}{lcccc}
\toprule
\makecell{Method} & \makecell{\# Bits (W/A)} & \makecell{ResNet-18} & \makecell{ResNet-50} & \makecell{MobileNetV2} \\
\midrule
Full-precision & $32 / 32$ & $69.76 / 89.08$ & $76.15 / 92.87$ & $71.88 / 90.29$\\
\midrule
B + AdaRound & $4 / 4$ & $68.32 / 88.13$ & $74.28 / 92.02$ & $28.46 / 52.60$  \\
B + FlexRound (Ours)& $4 / 4$ & $\mathbf{68.34} / \mathbf{88.19}$ & $74.42 / 92.04$ & $55.25 / 78.61$ \\ 
Q + AdaRound & $4 / 4$ & $68.19 / 88.18$ & $74.68 / 92.02$ & $56.68 / 80.95$  \\
Q + FlexRound (Ours)& $4 / 4$ & $68.23 / 88.22$ & $\mathbf{74.83} / \mathbf{92.11}$ & $\mathbf{61.56} / \mathbf{84.18}$ \\
\midrule
B + AdaRound & $3 / 3$ & $64.44 / 85.73$ & $68.80 / 88.79$ & $2.11 / 7.24$  \\
B + FlexRound (Ours)& $3 / 3$ & $64.61 / 85.85$ & $69.62 / 89.19$ & $8.80 / 21.79$ \\ 
Q + AdaRound & $3 / 3$ & $\mathbf{65.33} / \mathbf{86.60}$ & $71.80 / 90.72$ & $32.41 / 59.27$  \\
Q + FlexRound (Ours)& $3 / 3$ & $65.28 / 86.49$ & $\mathbf{71.84} / \mathbf{90.48}$ & $\mathbf{41.51} / \mathbf{68.02}$ \\
\bottomrule
\end{tabular}
\end{center}
\end{table}

% \iffalse

\newpage

\section{Cross-Layer Equalization and Absorbing High Biases as Preprocessing}

\begin{table}[h]
% \vskip -5pt
\caption{Top-1/Top-5 accuracy (\%) of MobileNetV2 with only weights quantized to $4$-bit on ImageNet. Here, the ``pre-trained model from BRECQ'' and ``pre-trained model from PyTorch'' columns show the results when using the pre-trained model provided from the BRECQ github repository and the official PyTorch repository, respectively. ``B $+$ X" denotes the implementation of X in the setting of BRECQ. ``Replacing ReLU6" indicates that every ReLU6 in MobileNetV2 is replaced by ReLU. ``CLE'' and ``AHB'' stand for cross-layer equalization and absorbing high biases, respectively.}
\label{tab:cle}
\begin{center}
\small
\begin{tabular}{lcc}
\toprule
\makecell{Method} & \makecell{pre-trained model \\ from BRECQ}  & \makecell{pre-trained model \\ from PyTorch} \\
\midrule
Full-precision & $72.62 / 90.67$ & $71.88 / 90.29$ \\
Replacing ReLU6 + CLE + AHB & $69.64 / 88.83$ & $71.53 / 90.19$ \\
\midrule
B + AdaRound & $69.46 / 88.85$ & $67.76 / 88.12$ \\
Replacing ReLU6 + CLE + AHB + B + AdaRound& $0.18 / 0.67$ & $\mathbf{70.03} / \mathbf{89.36}$ \\
\midrule
B + FlexRound & $\mathbf{70.82} / \mathbf{89.67}$ & $69.56 / 89.02$ \\ 
Replacing ReLU6 + CLE + AHB + B + FlexRound & $0.18 / 0.67$ & $69.44 / 89.00$ \\
\bottomrule
\end{tabular}
\end{center}
\end{table}

It is known that preprocessing pre-trained weights through cross-layer equalization (CLE) and absorbing high biases (AHB) exhibits a noticeable enhancement for the per-tensor quantization performance in vision models \citep{nagel2019data, nagel2021white}. To detect the effect of CLE and AHB on AdaRound and FlexRound as preprocessing, as seen in Table~\ref{tab:cle}, we also quantize the weights of MobileNetV2 preprocessed via CLE and AHB to $4$-bit using AdaRound and FlexRound in a linear symmetric per-tensor quantization format. %Following \citet{nagel2019data}, CLE and AHB are applied to MobileNetV2 after every ReLU6 in MobileNetV2 is replaced by ReLU. 
Following \citet{nagel2019data}, every ReLU6 in MobileNetV2 is replaced by ReLU when applying CLE and AHB to MobileNetV2.
When using the pre-trained model provided from the official PyTorch repository\footnote{\url{https://pytorch.org/vision/stable/models.html}}, utilizing CLE and AHB as preprocessing enhances the performance of `B + AdaRound' but not `B + FlexRound' so that `Replacing ReLU6 + CLE + AHB + B + AdaRound' shows better accuracy than `B + FlexRound' as well as `B + AdaRound'. In contrast, when using the pre-trained model provided from the BRECQ github repository\footnote{\url{https://github.com/yhhhli/BRECQ}}, utilizing CLE and AHB as preprocessing seriously hinders both `B + AdaRound' and `B + FlexRound' from performing well. %Depending on how a model is pre-trained, exploiting CLE and AHB as preprocessing can or cannot be effective, which means that the performance of `B + AdaRound + Replacing ReLU6 + CLE + AHB' can be very poor.
Depending on how a model is pre-trained, exploiting CLE and AHB as preprocessing can or cannot be effective. However, no matter which pre-trained model is chosen, `B + FlexRound' can consistently quantize weights well without any preprocessing, which implies that FlexRound would have its own advantages compared to other post-training weight quantization methods (that might need preprocessing for better performance).

\newpage

\section{Ablation Study on Sample Size}

\begin{figure}[H]
    % \vskip -12pt
    \centering
    \subfigure
    {
        \includegraphics[width=0.7\linewidth]{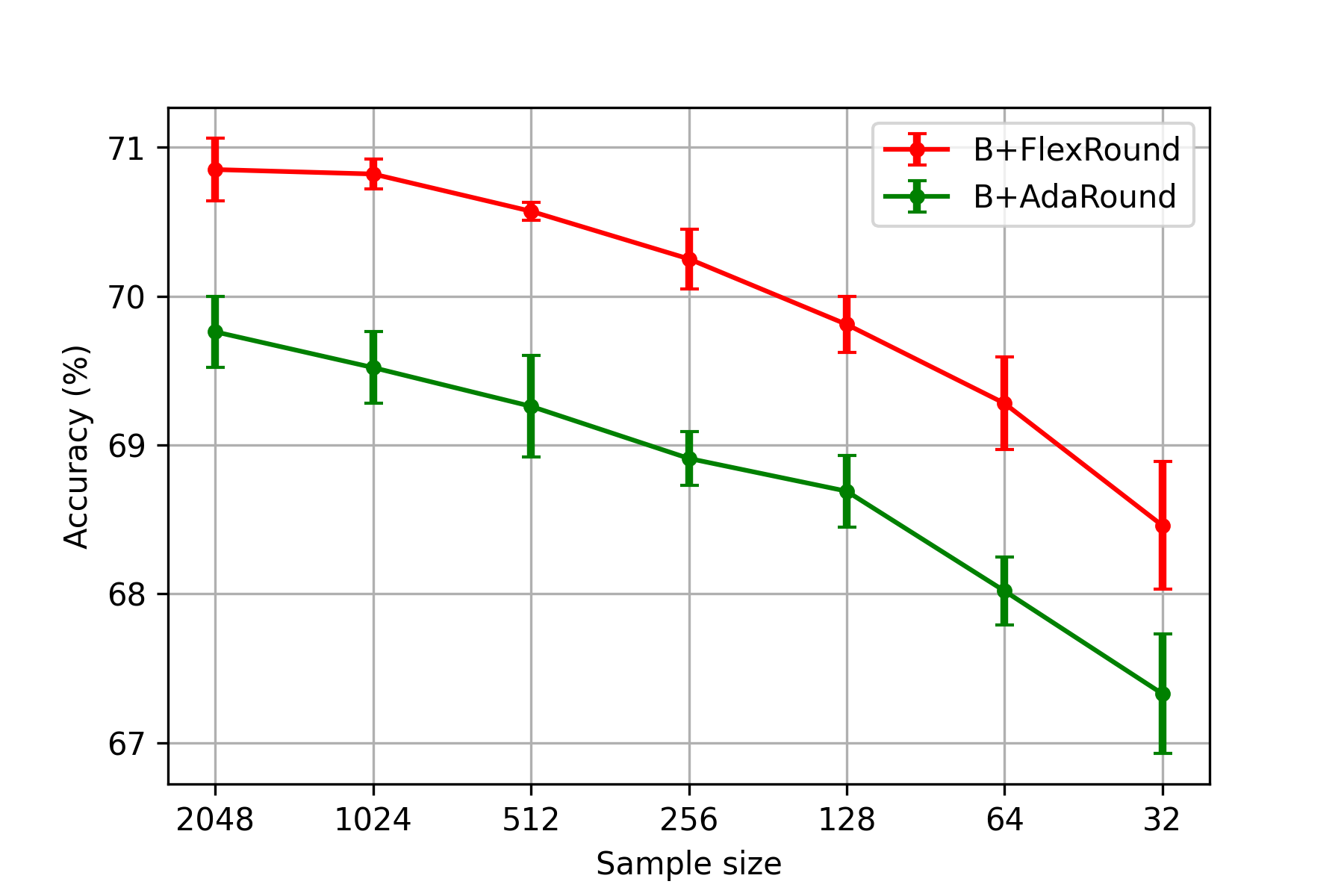}
        \label{fig:sample_size}
    }
    \caption{Ablation study on sample size when quantizing MobileNetV2 to $4$-bit. Only weights are quantized to $4$-bit, with activations kept in full-precision.}
    \label{fig:sample_size}
\end{figure}

No matter how much data is used, B+FlexRound always outperforms B+AdaRound. When the sample size decreases from 64 to 32, the accuracy of B+FlexRound declines by almost one percent. Correspondingly, a sample size of 32 would be a breakthrough point.

% fp_x = [2048,1024,512,256,128,64,32]
% ar = [68.03,67.77,67.36,66.71,66.26,65.43,63.93]
% ar_var = [0.13,0.21,0.25, 0.35,0.32,0.21, 0.48]
% fr = [69.85, 69.65, 69.41, 68.98, 68.60, 68.11, 67.17]
% fr_var = [0.12, 0.24, 0.12, 0.32,  0.30, 0.27, 0.31]
% from matplotlib import pyplot as plt
% x_tmp = range(len(fp_x))
% plt.errorbar(x_tmp, fr, fr_var, color = 'r',ecolor = 'r', marker='o', markerfacecolor= 'r', markersize= 4, capsize =3, elinewidth=2.5, label='B+FlexRound')
% plt.errorbar(x_tmp, ar, ar_var, color = 'g',ecolor = 'g', marker='o', markerfacecolor = 'g', markersize = 4, capsize = 3, elinewidth=2.5, label='B+AdaRound')
% plt.xlabel('Calibration set size')
% plt.ylabel('Accuracy')
% plt.grid()
% plt.legend()
% plt.xticks(range(len(fp_x)), fp_x)
% plt.show()

% \begin{table}[h]
% % \vskip -5pt
% \caption{\textcolor{blue}{Top-1/Top-5 accuracy (\%) on ImageNet by MobileNetV2 with only weights quantized into $4$-bit. ``B $+$ X" denotes the implementation of X in the setting of BRECQ.}}	
% \label{tab:ablation_sample}
% \begin{center}
% % \small
% \begin{tabular}{ccccccc}
% \Xhline{1pt}
% $2048$ & $1024$ & $512$ & $256$ & $128$ & $64$ & $32$ \\
% \hline
% $68.03 \pm 0.13$ & $67.77 \pm 0.21$ & $67.36 \pm 0.25$ & $66.71 \pm 0.35$ & $66.26 \pm 0.32$ & $65.43 \pm 0.21$ & $63.93 \pm 0.48$ \\
% $\mathbf{69.85 \pm 0.12}$ & $\mathbf{69.65 \pm 0.24}$ & $\mathbf{69.41 \pm 0.12}$ & $\mathbf{68.98 \pm 0.32}$ & $\mathbf{68.60 \pm 0.30}$ & $\mathbf{68.11 \pm 0.27}$ & $\mathbf{67.17 \pm 0.31}$ \\ 
% \Xhline{1pt}
% \end{tabular}
% \end{center}
% \end{table}

\newpage

\section{Combining an Additive Approach with a Division-based Approach}

\begin{table}[h]
\caption{Top-1/Top-5 accuracy (\%) on ImageNet when only weights are quantized. ``B $+$ X" expresses the implementation of X in the BRECQ's setting.}
\label{tab:imagenet_w_adaquant_flexround}
\begin{center}
\small
\begin{tabular}{lcccc}
\toprule
\makecell{Method} & \makecell{\# Bits (W/A)} & \makecell{ResNet-18} & \makecell{ResNet-50} & \makecell{MobileNetV2} \\
\midrule
Full-precision & $32 / 32$ & $71.00 / 89.97$ & $76.63 / 93.04$ & $72.62 / 90.67$\\
\midrule
B + AdaQuant & $4 / 32$ & $67.50 / 87.75$ & $72.79 / 90.77$ & $15.17 / 32.89$  \\
B + AdaQuant + FlexRound & $4 / 32$ & $69.81 / 89.21$ & $75.65 / 92.58$ & $70.15 / 89.34$ \\
B + FlexRound (Ours) & $4 / 32$ & $\mathbf{70.28} / \mathbf{89.44}$ & $\mathbf{75.95} / \mathbf{92.68}$ & $\mathbf{70.82} / \mathbf{89.67}$ \\
\midrule
B + AdaQuant & $3 / 32$ & $57.09 / 80.82$ & $52.13 / 75.22$ & $0.20 / 0.79$  \\
B + AdaQuant + FlexRound & $3 / 32$ & $67.93 / 88.08$ & $74.01 / 91.68$ & $65.58 / 86.63$ \\
B + FlexRound (Ours)& $3 / 32$ & $\mathbf{68.65} / \mathbf{88.54}$ & $\mathbf{74.38} / \mathbf{91.81}$ & $\mathbf{66.87} / \mathbf{87.56}$ \\
\midrule
B + AdaQuant & $2 / 32$ & $0.23 / 0.92$ & $0.10 / 0.50$ & $0.10 / 0.50$  \\
B + AdaQuant + FlexRound & $2 / 32$ & $61.13 / 83.93$ & $63.57 / 85.81$ & $44.56 / 71.25$ \\
B + FlexRound (Ours)& $2 / 32$ & $\mathbf{62.57} / \mathbf{84.84}$ & $\mathbf{63.67} / \mathbf{85.72}$ & $\mathbf{46.04} / \mathbf{72.48}$ \\
\bottomrule
\end{tabular}
\end{center}
\end{table}

One might wonder whether or not there comes any benefit from combining both element-wise addition and element-wise division. Although it would be interesting to combine AdaRound with FlexRound, such a combination would be challenging due to the fact that AdaRound cannot learn a quantization grid size, $s_1$ jointly with rounding. Alternatively, we combine AdaQuant with FlexRound. AdaQuant + FlexRound is superior to AdaQuant but inferior to FlexRound. This might be due to the naive combination of AdaQuant with FlexRound. Considering both element-wise addition and element-wise division would be an interesting future work.

% \fi

\newpage

\section{BERT on SQuAD}\label{appendix:squad}

\begin{table}[h]
\caption{F1 score for $\text{BERT}_{\text{Base}}$ and $\text{BERT}_{\text{Large}}$ on the SQuADv1 dataset when both weights and activations are quantized to $8$-bit. ``Q $+$ X" represent the implementation of X in the QDrop's setting.}	
\label{tab:squad}
\begin{center}
\small
\begin{tabular}{lccc}
\toprule
\makecell{Method} & \makecell{\# Bits (W/A)} & \makecell{$\text{BERT}_{\text{Base}}$} & \makecell{$\text{BERT}_{\text{Large}}$ }  \\
\midrule
Full-precision & $32 / 32$ & $87.05$ & $89.31$ \\
\midrule
Q + AdaRound & $8 / 8$ & $86.90$ & $88.89$  \\
Q + FlexRound (Ours)& $8 / 8$ & $\mathbf{87.25}$ & $\mathbf{89.25}$  \\
\bottomrule
\end{tabular}
\end{center}
\end{table}

\begin{table}[h]
\caption{Hyper-parameter selection for fine-tuning $\text{BERT}_{\text{Base}}$ and $\text{BERT}_{\text{Large}}$ on the SQuADv1 dataset.}	
\begin{center}
\small
\begin{tabular}{ccccc}
\toprule
\makecell{Learning rate} & \makecell{Batch size} & \makecell{Epoch} & \makecell{Maximum sequence length} & \makecell{Document stride}   \\
\midrule
$1$e-$4$ & $32$ & $4$ & $384$ & $128$ \\
\bottomrule
\label{tab:squad_finetune}
\end{tabular}
\end{center}
\end{table}

Table~\ref{tab:squad} shows the performace of FlexRound on the SQuADv1 \citep{2016arXiv160605250R} dataset\footnote{\url{https://huggingface.co/datasets/squad}} for the BERT models. Both $\text{BERT}_{\text{Base}}$ and $\text{BERT}_{\text{Large}}$ are uncased models. For reconstruction, we select $1024$ samples from the training dataset of SQuADv1 at random without any modification. For `Q + FlexRound', the learning rate is set to $1$e-$4$ for both models. For both `Q + AdaRound' and `Q + FlexRound', the batch size and the number of iterations for reconstruction are $64$ and $20k$ respectively. We use the Adam optimizer for all methods and models. The other experimental setting of `Q + AdaRound' follows \citet{wei2022qdrop}. 

Table~\ref{tab:squad_finetune} shows the hyper-parameter selection for fine-tuning. The same configuration is used for both $\text{BERT}_{\text{Base}}$ and $\text{BERT}_{\text{Large}}$. The other setting for fine-tuning and the evaluation method are the same as HuggingFace repository\footnote{\url{https://github.com/huggingface/transformers/tree/main/examples/pytorch/question-answering}}.

\newpage

\section{BERT and GPT-Neo on GLUE}\label{appendix:nlu}

\begin{table}[h]
\caption{Hyper-parameter selection for fine-tuning $\text{BERT}_{\text{Base}}$, $\text{BERT}_{\text{Large}}$, $\text{GPT-Neo}_{125\text{M}}$, $\text{GPT-Neo}_{1.3\text{B}}$, and $\text{GPT-Neo}_{2.7\text{B}}$ on GLUE.}
\label{tab:glue-ft}
\begin{center}
\small
\begin{tabular}{cccccc}
\toprule
\makecell{Configuration} & $\text{BERT}_{\text{Base}}$ & $\text{BERT}_{\text{Large}}$ & $\text{GPT-Neo}_{125\text{M}}$ & $\text{GPT-Neo}_{1.3\text{B}}$ & $\text{GPT-Neo}_{2.7\text{B}}$ \\ 
\midrule
Learning Rate& $2$e-$5$ & $2$e-$5$ & $2$e-$5$ & $2$e-$5$ & $1$e-$5$\\
 Batch Size & $32$ & $32$ & $32$ & $32$ & $16$ \\
Epoch &&&   $3$ \\
Maximum Sequence Length &&& $128$ \\
Weight Decay &&& $0.01$ \\
\bottomrule
\end{tabular}
\end{center}
\vskip -3pt
\end{table}

\begin{table*}[h]
\caption{Performance of BERT and GPT-Neo fine-tuned on GLUE. For evaluation, matched and mismatched accuracies are reported for MNLI, F1 score and accuracy are reported for QQP, Mathews correlation is reported for CoLA, Pearson and Spearman correlations are reported for STS-B, and accuracy is reported for the others. ``Q $+$ X" indicates the implementation of X in the QDrop's setting.} 
\label{tab:glue_appendix}
\begin{center}
\small
% \resizebox{0.9\linewidth}{!}{
\begin{tabular}{clccccc}
\toprule
Dataset & \makecell{Method} & $\text{BERT}_{\text{BASE}}$ & $\text{BERT}_{\text{LARGE}}$ & $\text{GPT-Neo}_{125\text{M}}$ & $\text{GPT-Neo}_{1.3\text{B}}$ & $\text{GPT-Neo}_{2.7\text{B}}$ \\
% Method & \makecell{\# Bits \\ (W./A.)} & \makecell{MNLI \\ (acc. m/mm)} & \makecell{QQP \\ (F1/acc.)} & \makecell{QNLI \\ (acc.)} & \makecell{SST-2 \\ (acc.)} & \makecell{CoLA \\ (Matthews corr.)} & \makecell{STS-B \\ (Pearson/Spearman corr.)} & \makecell{MRPC \\ (acc.)} & \makecell{RTE \\ (acc.)} \\
\midrule
 & Full-precision & $84.49 / 85.20$ & $86.05 / 85.98$ & $79.11 / 79.63$ & $85.12 / 86.04$ & $86.36 / 87.02$ \\
\cmidrule{2-7}
MNLI  & Q+AdaRound & $83.69 / 84.61$ & $85.75 / 85.86$ & $72.67 / 74.11$ & $84.90 / 85.82$ & $86.33 / 86.75 $ \\
& Q+FlexRound (Ours)& $\mathbf{84.53} / \mathbf{84.98}$ & $\mathbf{85.93} / \mathbf{85.99}$ & $\mathbf{72.94} / \mathbf{74.24}$ & $\mathbf{85.56} / \mathbf{86.14}$ & $\mathbf{86.41} / \mathbf{86.89}$ \\
\midrule
& Full-precision & $88.06 / 91.08$ & $88.66 / 91.59$ & $85.20 / 88.99$ & $88.26 / 91.28$ & $88.62 / 91.50$ \\
\cmidrule{2-7}
QQP & Q+AdaRound & $87.65 / 90.58$ & $87.48 / 90.62$ & $72.97 / 79.35$ & $87.98 / 91.04$ & $88.38 / 91.27$ \\
& Q+FlexRound (Ours)& $\mathbf{87.81} / \mathbf{90.83}$ & $\mathbf{88.38} / \mathbf{91.31}$ & $\mathbf{73.75} / \mathbf{80.65}$ & $\mathbf{88.27} / \mathbf{91.18}$ & $\mathbf{88.60} / \mathbf{91.39}$ \\
\midrule
& Full-precision & $91.25$ & $92.13$ & $85.15$ & $91.36$ & $92.46$ \\
\cmidrule{2-7}
QNLI & Q+AdaRound & $91.16$ & $\mathbf{92.24}$ & $\mathbf{80.87}$ & $91.40$ & $92.04$ \\
& Q+FlexRound (Ours)& $\mathbf{91.16}$ & $92.04$ & $80.52$ & $\mathbf{91.54}$ & $\mathbf{92.50}$ \\
\midrule
& Full-precision & $93.00$ & $92.78$ & $89.91$ & $93.35$ & $94.50$ \\
\cmidrule{2-7}
SST-2 & Q+AdaRound & $\mathbf{92.66}$ & $93.00$ & $\mathbf{84.75}$ & $92.55$ & $93.81$ \\
& Q+FlexRound (Ours)& $92.43$ & $\mathbf{93.58}$ & $83.03$ & $\mathbf{93.12}$ & $\mathbf{94.04}$ \\
\midrule
& Full-precision & $58.55$ & $63.57$ & $37.83$ & $57.42$ & $58.88$ \\ 
\cmidrule{2-7}
CoLA & Q+AdaRound & $56.79$ & $54.30$ & $20.15$ & $58.93$ &  $57.14$ \\
& Q+FlexRound (Ours)& $\mathbf{57.53}$ & $\mathbf{60.57}$ & $\mathbf{21.59}$ & $\mathbf{59.30}$ & $\mathbf{57.37}$  \\
\midrule
& Full-precision & $88.52 / 88.20$ & $88.98 / 88.89$ & $79.87 / 80.12$ & $88.94 / 88.90$ & $89.75 / 89.82$ \\ 
\cmidrule{2-7}
STS-B & Q+AdaRound & $88.00 / 87.53$ & $86.87 / 86.69$ & $\mathbf{68.55} / 68.25$ & $\mathbf{88.97} / \mathbf{88.77}$ & $89.03 / \mathbf{88.91}$ \\
& Q+FlexRound (Ours)& $\mathbf{88.29} / \mathbf{87.91}$ & $\mathbf{88.82} / \mathbf{88.76}$ & $67.65 / \mathbf{68.34}$ & $88.82 / 88.58$ & $\mathbf{89.06} / 88.69$ \\
\midrule
& Full-precision & $85.05$ & $85.54$ & $80.15$ & $85.05$ & $87.99$ \\ 
\cmidrule{2-7}
MRPC & Q+AdaRound & $81.62$ & $82.35$ & $75.25$ & $84.80$ &  $85.78$ \\
& Q+FlexRound (Ours)& $\mathbf{84.07}$ & $\mathbf{84.31}$ & $\mathbf{75.49}$ & $\mathbf{85.05}$ & $\mathbf{86.76}$ \\
\midrule
& Full-precision & $64.62$ & $71.19$ & $64.98$ & $76.17$ & $80.87$ \\ 
\cmidrule{2-7}
RTE & Q+AdaRound & $63.54$ & $66.79$ & $62.82$ & $75.09$ & $80.51$ \\
& Q+FlexRound (Ours)& $\mathbf{64.62}$ & $\mathbf{68.95}$ & $\mathbf{62.82}$ & $\mathbf{76.17}$ & $\mathbf{81.23}$ \\
\bottomrule
\end{tabular}
% }
\end{center}
% \vskip -10pt
\end{table*}

% \clearpage
% 
% \begin{table}[h]
% \caption{Hyper-parameter selection for fine-tuning $\text{BERT}_{\text{Base}}$, $\text{BERT}_{\text{Large}}$, $\text{GPT-Neo}_{125\text{M}}$, $\text{GPT-Neo}_{1.3\text{B}}$, and $\text{GPT-Neo}_{2.7\text{B}}$ on GLUE.}
% \label{tab:glue-ft}
% \begin{center}
% \small
% \begin{tabular}{cccccc}
% \toprule
% \makecell{Configuration} & $\text{BERT}_{\text{Base}}$ & $\text{BERT}_{\text{Large}}$ & $\text{GPT-Neo}_{125\text{M}}$ & $\text{GPT-Neo}_{1.3\text{B}}$ & $\text{GPT-Neo}_{2.7\text{B}}$ \\ 
% \midrule
% Learning Rate& $2$e-$5$ & $2$e-$5$ & $2$e-$5$ & $2$e-$5$ & $1$e-$5$\\
%  Batch Size & $32$ & $32$ & $32$ & $32$ & $16$ \\
% Epoch &&&   $3$ \\
% Maximum Sequence Length &&& $128$ \\
% Weight Decay &&& $0.01$ \\
% \bottomrule
% \end{tabular}
% \end{center}
% \end{table}

To investigate the natural language understanding performance of FlexRound for BERT\footnote{\url{https://huggingface.co/bert-base-uncased}} to GPT-Neo\footnote{\url{https://huggingface.co/EleutherAI/gpt-neo-1.3B}}, we directly fine-tune pre-trained models on the GLUE\footnote{\url{https://huggingface.co/datasets/glue}} benchmark. For BERT, we use uncased models. Hyper-parameter selection for fine-tuning a pre-trained model is given in Table~\ref{tab:glue-ft}.  We use the Huggingface repository\footnote{\url{https://github.com/huggingface/transformers/tree/main/examples/pytorch/text-classification}} for fine-tuning without any modification.

In Table \ref{tab:glue_appendix}, for reconstruction, we randomly sample $1024$ examples from the training dataset without any modification. For all experiments, the batch size is 64, and the maximum sequence length of all experiments is 128. We use the Adam optimizer for all methods and models. In the QDrop's setting, the probability of dropping activation quantization is set to $0.5$. The experimental setting of `Q + AdaRound' follows \citet{wei2022qdrop}. We also utilize the Huggingface repository\footnote{\url{https://github.com/huggingface/transformers/tree/main/examples/pytorch/text-classification}} for the evaluation method without any modification.

For some datasets (QNLI, SST-2, and STS-B), `Q + FlexRound' does not outperform `Q + AdaRound' as shown in Table \ref{tab:glue_appendix}. This suggests that there may be certain tasks where FlexRound has room for improvement. However, this outcome is due to the fact that the learning rate for $s_1$, $\mS_2$, and $\vs_3$ is set to $2$e-$4$ for BERT and $3$e-$4$ for GPT-Neo to demonstrate that `Q + FlexRound' can broadly surpass `Q + AdaRound' without the need of significant efforts to select the optimal learning rate for each task. When the learning rate is fine-tuned for the datasets where `Q + FlexRound' falls short of `Q + AdaRound', we can observe that `Q + FlexRound' outperforms `Q + AdaRound' in most cases, as depicted in the table below.

\begin{table*}[h]
\caption{Performance of BERT and GPT-Neo fine-tuned on GLUE after tuning the learning rate of $s_1$, $\mS_2$, and $\vs_3$ for the tasks where `Q + FlexRound' falls short of `Q + AdaRound'. Pearson and Spearman correlations are reported for STS-B, and accuracy is reported for the others. ``Q $+$ X" indicates the implementation of X in the QDrop's setting.} 
\label{tab:glue_appendix2}
\begin{center}
\small
% \resizebox{0.9\linewidth}{!}{
\begin{tabular}{clccccc}
\toprule
Dataset & \makecell{Method} & $\text{BERT}_{\text{BASE}}$ & $\text{BERT}_{\text{LARGE}}$ & $\text{GPT-Neo}_{125\text{M}}$ & $\text{GPT-Neo}_{1.3\text{B}}$ & $\text{GPT-Neo}_{2.7\text{B}}$ \\
% Method & \makecell{\# Bits \\ (W./A.)} & \makecell{MNLI \\ (acc. m/mm)} & \makecell{QQP \\ (F1/acc.)} & \makecell{QNLI \\ (acc.)} & \makecell{SST-2 \\ (acc.)} & \makecell{CoLA \\ (Matthews corr.)} & \makecell{STS-B \\ (Pearson/Spearman corr.)} & \makecell{MRPC \\ (acc.)} & \makecell{RTE \\ (acc.)} \\
\midrule
& Full-precision & $91.25$ & $92.13$ & $85.15$ & $91.36$ & $92.46$ \\
\cmidrule{2-7}
QNLI & Q+AdaRound & $91.16$ & $92.24$ & $80.87$ & $91.40$ & $92.04$ \\
& Q+FlexRound (Ours)& $\mathbf{91.16}$ & $\mathbf{92.26}$ & $\mathbf{82.72}$ & $\mathbf{91.54}$ & $\mathbf{92.50}$ \\
\midrule
& Full-precision & $93.00$ & $92.78$ & $89.91$ & $93.35$ & $94.50$ \\
\cmidrule{2-7}
SST-2 & Q+AdaRound & $92.66$ & $93.00$ & $\mathbf{84.75}$ & $92.55$ & $93.81$ \\
& Q+FlexRound (Ours)& $\mathbf{92.66}$ & $\mathbf{93.58}$ & $83.72$ & $\mathbf{93.12}$ & $\mathbf{94.04}$ \\
\midrule
& Full-precision & $88.52 / 88.20$ & $88.98 / 88.89$ & $79.87 / 80.12$ & $88.94 / 88.90$ & $89.75 / 89.82$ \\ 
\cmidrule{2-7}
STS-B & Q+AdaRound & $88.00 / 87.53$ & $86.87 / 86.69$ & $68.55 / 68.25$ & $88.97 / 88.77$ & $89.03 / 88.91$ \\
& Q+FlexRound (Ours)& $\mathbf{88.29} / \mathbf{87.91}$ & $\mathbf{88.82} / \mathbf{88.76}$ & $\mathbf{69.25} / \mathbf{69.58}$ & $\mathbf{89.20} / \mathbf{88.99}$ & $\mathbf{89.06} / \mathbf{88.96}$ \\

\bottomrule
\end{tabular}
% }
\end{center}
% \vskip -10pt
\end{table*}

\newpage

\section{GPT-Neo and OPT on WikiText2 and PTB}\label{appendix:nlg_finetune}

\begin{table}[h]
\caption{Hyper-parameter selection for fine-tuning $\text{GPT-Neo}_{125\text{M}}$, $\text{GPT-Neo}_{1.3\text{B}}$, $\text{GPT-Neo}_{2.7\text{B}}$, $\text{OPT}_{125\text{M}}$, $\text{OPT}_{1.3\text{B}}$, and $\text{OPT}_{2.7\text{B}}$ on the WikiText2 and PTB datasets.} \label{tab:hyperparameter_nlg}
\begin{center}
\begin{tabular}{clcccccc}
\toprule
Dataset & \makecell{Configuration} & $\text{GPT-Neo}_{125\text{M}}$ & $\text{GPT-Neo}_{1.3\text{B}}$ & $\text{GPT-Neo}_{2.7\text{B}}$ & $\text{OPT}_{125\text{M}}$ & $\text{OPT}_{1.3\text{B}}$ & $\text{OPT}_{2.7\text{B}}$ \\
\midrule
WikiText2 & Learning rate & $3$e-$5$ & $4$e-$6$ & $1$e-$6$ & $3$e-$5$ & $3$e-$6$ & $2$e-$6$ \\
& Batch size & $8$ & $4$ & $2$ & $8$ & $4$ & $2$ \\
\midrule
PTB & Learning rate & $9$e-$5$ & $1$e-$5$ & $6$e-$6$ & $1$e-$5$ & $9$e-$6$ & $6$e-$6$ \\
& Batch size & $8$ & $4$ & $2$ & $8$ & $4$ & $2$ \\
\bottomrule
\end{tabular}
\end{center}
\end{table}

\begin{table}[h]
\caption{Hyper-parameter selection for `Q + FlexRound' in Table~\ref{tab:clm_finetuned}. For all experiments, the sample size and the number of iterations are set to $128$ and $500$, respectively.}\label{tab:hyperparameter_nlg_finetuned}
\begin{center}
\begin{tabular}{clcccccc}
\toprule
Dataset & \makecell{Configuration} & $\text{GPT-Neo}_{125\text{M}}$ & $\text{GPT-Neo}_{1.3\text{B}}$ & $\text{GPT-Neo}_{2.7\text{B}}$ & $\text{OPT}_{125\text{M}}$ & $\text{OPT}_{1.3\text{B}}$ & $\text{OPT}_{2.7\text{B}}$ \\
\midrule
WikiText2 & Learning rate & $5$e-$3$ & $4$e-$4$ & $4$e-$3$ & $3$e-$5$ & $7$e-$6$ & $1$e-$5$ \\
& Batch size & $32$ & $16$ & $8$ & $32$ & $16$ & $8$ \\
\midrule
PTB & Learning rate & $5$e-$3$ & $7$e-$3$ & $7$e-$3$ & $5$e-$5$ & $3$e-$5$ & $8$e-$6$ \\
& Batch size & $32$ & $16$ & $8$ & $32$ & $16$ & $8$ \\
\bottomrule
\end{tabular}
\end{center}
\end{table}

To evaluate FlexRound for natural language generation tasks, we utilize GPT-Neo\footnote{\url{https://huggingface.co/EleutherAI/gpt-neo-1.3B}} and OPT\footnote{\url{https://huggingface.co/facebook/opt-1.3b}} fine-tuned on the WikiText2 \footnote{\url{https://huggingface.co/datasets/wikitext}} and PTB \footnote{\url{https://huggingface.co/datasets/ptb_text_only}} datasets for $10$ epochs. Table~\ref{tab:hyperparameter_nlg} reports hyper-parameter selection for fine-tuning a pre-trained language model. We utilize the Huggingface repository\footnote{\url{https://github.com/huggingface/transformers/tree/main/examples/pytorch/language-modeling}} for fine-tuning without any modification.

For reconstruction, We extract $128$ random samples from the training dataset without any modification, and the number of iterations is fixed to $500$. We use the Adam optimizer for all methods and models. The learning rate and batch size for `Q + FlexRound' in Table~\ref{tab:clm_finetuned} are shown in Table~\ref{tab:hyperparameter_nlg_finetuned}. 
The batch size of `Q + AdaRound' is same as the batch size of `Q + FlexRound'. The other experimental setting of `Q + AdaRound' follows \citet{wei2022qdrop}. The probability of dropping activation quantization is set to $0.5$ in the QDrop's setting. We also use the Huggingface repository\footnote{\url{https://github.com/huggingface/transformers/tree/main/examples/pytorch/language-modeling}} for the evaluation method without any modification.
% Other settings are the same as Appendix~\ref{appendix:nlg}.}

\newpage

\section{GPT-2 on WebNLG}\label{appendix:webnlg}

In Table~\ref{tab:lora}, we utilize the GPT-2 models and the WebNLG dataset available from the LoRA repository\footnote{\url{https://github.com/microsoft/LoRA}}. Namely, all LoRA checkpoints are loaded from the repository and merged to GPT-2. For reconstruction in all experiments, we use $128$ random samples from the training dataset of WebNLG without any modification, and the number of iterations and the batch size are set to $500$ and $8$ respectively. For `Q + FlexRound', the learning rate is set to $5$e-$3$ for GPT-2 medium and $3$e-$3$ for GPT-2 large, respectively. The other experimental setup of `Q + AdaRound' follows \citet{wei2022qdrop}.

\newpage

\section{LLaMA on Common Sense Reasoning and WikiText2}\label{appendix:llama}

\begin{table*}[h]
% \vskip -0.1in
\caption{Zero-shot performance of LLaMA-$7$B, LLaMA-$13$B, and LLaMA-$33$B on $6$ common sense reasoning benchmarks (BoolQ, PIQA, HellaSwag, WinoGrande, ARC easy and challenge, and OBQA) and the causal language modeling task on WikiText$2$. The accuracy ($\%$) and the perplexity (PPL) are reported for common sense reasoning tasks and the causal language modeling task, respectively. The lower PPL, the better. ``Q $+$ X" implies the implementation of X in the QDrop's setting. The weights of attention and feed-forward sub-layers are quantized to $8$-bit in a per-channel asymmetric format, whereas the input activations of those sub-layers are quantized to $8$-bit in a per-tensor asymmetric scheme.}
\label{tab:plm_llama_zero_appendix}
\begin{center}
\small
\resizebox{\linewidth}{!}{
\begin{tabular}{clccccccccc}
\toprule
Model & \makecell{Method}  & \# Bits (W/A) & BoolQ & PIQA & HellaSwag &  WinoGrande & ARC-e & ARC-c & OBQA & WikiText2\\
\midrule
& Half-precision & $16 / 16$ & $73.15$ & $77.31$ & $72.96$ & $67.09$ & $52.48$ & $41.38$ & $42.40$ & $8.90$ \\
\cmidrule{2-11}
LLaMA-$7$B & Q+AdaRound & $8 / 8$ & $70.12$ & $75.08$ & $69.89$ & $65.82$ & $51.47$ & $39.42$ & $39.00$ & $10.38$ \\
& Q+FlexRound (Ours) & $8 / 8$ & $\mathbf{73.76}$ & $\mathbf{76.66}$ & $\mathbf{71.75}$ & $\mathbf{67.01}$ & $\mathbf{52.31}$ & $\mathbf{40.02}$ & $\mathbf{42.20}$ & $\mathbf{9.25}$ \\
\midrule
& Half-precision & $16 / 16$ & $68.53$ & $79.11$ & $76.23$ & $70.01$ & $59.89$ & $44.54$ & $42.20$ & $7.73$ \\
\cmidrule{2-11}
LLaMA-$13$B & Q+AdaRound & $8 / 8$ & $66.09$ & $76.44$ & $72.06$ & $66.30$ & $57.32$ & $43.00$ & $39.60$ & $9.07$ \\
& Q+FlexRound (Ours) & $8 / 8$ & $\mathbf{68.59}$ & $\mathbf{78.67}$ & $\mathbf{75.21}$ & $\mathbf{70.64}$ & $\mathbf{58.88}$ & $\mathbf{43.60}$ & $\mathbf{41.20}$ & $\mathbf{8.01}$ \\
\midrule
& Half-precision & $16 / 16$ & $68.38$ & $80.09$ & $79.21$ & $72.93$ & $58.92$ & $45.48$ & $42.00$ & $6.35$ \\
\cmidrule{2-11}
LLaMA-$33$B & Q+AdaRound & $8 / 8$ & $64.86$ & $74.65$ & $68.64$ & $57.93$ & $49.28$ & $36.95$ & $41.00$ & $10.39$ \\
& Q+FlexRound (Ours) & $8 / 8$ & $\mathbf{69.08}$ & $\mathbf{79.16}$ & $\mathbf{77.43}$ & $\mathbf{72.53}$ & $\mathbf{56.61}$ & $\mathbf{44.97}$ & $\mathbf{44.00}$ & $\mathbf{6.82}$ \\
% \midrule
% & Half-precision & $85.96$ & $82.48$ & $82.20$ & $80.03$ & $74.87$ & $56.23$ & $47.00$ & \\
% \cmidrule{2-10}
% Five-shot  & Q+AdaRound & $85.96$ & $82.48$ & $82.20$ & $80.03$ & $74.87$ & $56.23$ & $47.00$ & \\
% & Q+FlexRound (Ours)& $\mathbf{85.32}$ & $\mathbf{80.90}$ & $\mathbf{80.52}$ & $\mathbf{78.37}$ & $\mathbf{71.72}$ & $\mathbf{53.16}$ & $\mathbf{46.80}$ & \\
\bottomrule
\end{tabular}
}
\end{center}
% \vskip -0.1in
\end{table*}

\begin{table*}[h]
% \vskip -0.1in
\caption{Five-shot performance of LLaMA-$7$B, LLaMA-$13$B, and LLaMA-$33$B on $6$ common sense reasoning benchmarks (BoolQ, PIQA, HellaSwag, WinoGrande, ARC easy and challenge, and OBQA). The accuracy ($\%$) is reported for common sense reasoning tasks. ``Q $+$ X" implies the implementation of X in the QDrop's setting. The weights of attention and feed-forward sub-layers are quantized to $8$-bit in a per-channel asymmetric format, whereas the input activations of those sub-layers are quantized to $8$-bit in a per-tensor asymmetric scheme.} 
\label{tab:plm_llama_five_appendix}
\begin{center}
\small
\resizebox{\linewidth}{!}{
\begin{tabular}{clcccccccc}
\toprule
Model & \makecell{Method} & \# Bits (W/A) & BoolQ & PIQA & HellaSwag &  WinoGrande & ARC-e & ARC-c & OBQA \\
\midrule
& Half-precision & $16 / 16$ & $76.33$ & $79.38$ & $75.35$ & $69.69$ & $65.78$ & $45.56$ & $44.00$ \\
\cmidrule{2-10}
LLaMA-$7$B & Q+AdaRound & $8 / 8$ & $68.38$ & $76.55$ & $72.60$ & $\mathbf{70.40}$ & $62.75$ & $44.45$ & $42.20$ \\
& Q+FlexRound (Ours) & $8 / 8$ & $\mathbf{76.76}$ & $\mathbf{78.07}$ & $\mathbf{74.17}$ & $69.14$ & $\mathbf{64.14}$ & $\mathbf{45.05}$ & $\mathbf{43.60}$ \\
\midrule
& Half-precision & $16 / 16$ & $81.90$ & $79.98$ & $78.41$ & $75.61$ & $70.79$ & $50.43$ & $47.20$ \\
\cmidrule{2-10}
LLaMA-$13$B & Q+AdaRound & $8 / 8$ & $67.95$ & $77.80$ & $74.32$ & $73.01$ & $64.52$ & $45.82$ & $44.40$ \\
& Q+FlexRound (Ours) & $8 / 8$ & $\mathbf{78.29}$ & $\mathbf{80.20}$ & $\mathbf{77.26}$ & $\mathbf{75.37}$ & $\mathbf{67.68}$ & $\mathbf{49.32}$ & $\mathbf{46.40}$ \\
\midrule
& Half-precision & $16 / 16$ & $85.96$ & $82.48$ & $82.20$ & $80.03$ & $74.87$ & $56.23$ & $47.00$ \\
\cmidrule{2-10}
LLaMA-$33$B& Q+AdaRound  & $8 / 8$ & $68.38$ & $80.09$ & $79.21$ & $72.93$ & $58.92$ & $45.48$ & $42.00$ \\
& Q+FlexRound (Ours) & $8 / 8$ & $\mathbf{85.32}$ & $\mathbf{80.90}$ & $\mathbf{80.52}$ & $\mathbf{78.37}$ & $\mathbf{71.72}$ & $\mathbf{53.16}$ & $\mathbf{46.80}$ \\
% \midrule
% & Half-precision & $85.96$ & $82.48$ & $82.20$ & $80.03$ & $74.87$ & $56.23$ & $47.00$ & \\
% \cmidrule{2-10}
% Five-shot  & Q+AdaRound & $85.96$ & $82.48$ & $82.20$ & $80.03$ & $74.87$ & $56.23$ & $47.00$ & \\
% & Q+FlexRound (Ours)& $\mathbf{85.32}$ & $\mathbf{80.90}$ & $\mathbf{80.52}$ & $\mathbf{78.37}$ & $\mathbf{71.72}$ & $\mathbf{53.16}$ & $\mathbf{46.80}$ & \\
\bottomrule
\end{tabular}
}
\end{center}
% \vskip -0.1in
\end{table*}

\clearpage

\begin{table*}[h]
% \vskip -0.1in
\caption{Zero-shot performance of LLaMA-$7$B, LLaMA-$13$B, and LLaMA-$33$B on $6$ common sense reasoning benchmarks (BoolQ, PIQA, HellaSwag, WinoGrande, ARC easy and challenge, and OBQA) and the causal language modeling task on WikiText$2$. The accuracy ($\%$) and the perplexity (PPL) are reported for common sense reasoning tasks and the causal language modeling task, respectively. The lower PPL, the better. ``B $+$ X" implies the implementation of X in the BRECQ's setting. The weights of attention and feed-forward sub-layers are quantized to $4$-bit in a per-channel asymmetric format, whereas the input activations of those sub-layers are kept in half-precision.}
\label{tab:plm_llama_zero_appendix_w4a16}
\begin{center}
\small
\resizebox{\linewidth}{!}{
\begin{tabular}{clccccccccc}
\toprule
Model & \makecell{Method}  & \# Bits (W/A) & BoolQ & PIQA & HellaSwag &  WinoGrande & ARC-e & ARC-c & OBQA & WikiText2\\
\midrule
& Half-precision & $16 / 16$ & $73.15$ & $77.31$ & $72.96$ & $67.09$ & $52.48$ & $41.38$ & $42.40$ & $8.90$ \\
\cmidrule{2-11}
LLaMA-$7$B & B+AdaRound & $4 / 16$ & $70.46$ & $77.04$ & $71.73$ & $\mathbf{68.27}$ & $\mathbf{51.73}$ & $\mathbf{40.44}$ & $42.00$ & $9.69$ \\
& B+FlexRound (Ours) & $4 / 16$ & $\mathbf{70.73}$ & $\mathbf{77.75}$ & $\mathbf{71.97}$ & $66.06$ & $50.80$ & $40.27$ & $\mathbf{42.20}$ & $\mathbf{9.18}$ \\
\midrule
& Half-precision & $16 / 16$ & $68.53$ & $79.11$ & $76.23$ & $70.01$ & $59.89$ & $44.54$ & $42.20$ & $7.73$ \\
\cmidrule{2-11}
LLaMA-$13$B & B+AdaRound & $4 / 16$ & $\mathbf{67.55}$ & $\mathbf{78.94}$ & $75.50$ & $69.85$ & $58.42$ & $43.00$ & $\mathbf{43.40}$ & $8.07$ \\
& B+FlexRound (Ours) & $4 / 16$ & $66.39$ & $78.78$ & $\mathbf{75.52}$ & $\mathbf{70.40}$ & $\mathbf{59.55}$ & $\mathbf{43.77}$ & $42.80$ & $\mathbf{7.90}$ \\
\midrule
& Half-precision & $16 / 16$ & $68.38$ & $80.09$ & $79.21$ & $72.93$ & $58.92$ & $45.48$ & $42.00$ & $6.35$ \\
\cmidrule{2-11}
LLaMA-$33$B & B+AdaRound & $4 / 16$ & $\mathbf{69.39}$ & $79.27$ & $77.77$ & $72.69$ & $57.03$ & $44.62$ & $43.00$ & $6.88$ \\
& B+FlexRound (Ours) & $4 / 16$ & $67.19$ & $\mathbf{80.25}$ & $\mathbf{79.01}$ & $\mathbf{72.61}$ & $\mathbf{57.79}$ & $\mathbf{44.88}$ & $\mathbf{43.80}$ & $\mathbf{6.63}$ \\
% \midrule
% & Half-precision & $85.96$ & $82.48$ & $82.20$ & $80.03$ & $74.87$ & $56.23$ & $47.00$ & \\
% \cmidrule{2-10}
% Five-shot  & Q+AdaRound & $85.96$ & $82.48$ & $82.20$ & $80.03$ & $74.87$ & $56.23$ & $47.00$ & \\
% & Q+FlexRound (Ours)& $\mathbf{85.32}$ & $\mathbf{80.90}$ & $\mathbf{80.52}$ & $\mathbf{78.37}$ & $\mathbf{71.72}$ & $\mathbf{53.16}$ & $\mathbf{46.80}$ & \\
\bottomrule
\end{tabular}
}
\end{center}
% \vskip -0.1in
\end{table*}

\begin{table*}[h]
% \vskip -0.1in
\caption{Five-shot performance of LLaMA-$7$B, LLaMA-$13$B, and LLaMA-$33$B on $6$ common sense reasoning benchmarks (BoolQ, PIQA, HellaSwag, WinoGrande, ARC easy and challenge, and OBQA). The accuracy ($\%$) is reported for common sense reasoning tasks. ``B $+$ X" implies the implementation of X in the BRECQ's setting. The weights of attention and feed-forward sub-layers are quantized to $4$-bit in a per-channel asymmetric format, whereas the input activations of those sub-layers are kept in half-precision.} 
\label{tab:plm_llama_five_appendix_w4a16}
\begin{center}
\small
\resizebox{\linewidth}{!}{
\begin{tabular}{clcccccccc}
\toprule
Model & \makecell{Method} & \# Bits (W/A) & BoolQ & PIQA & HellaSwag &  WinoGrande & ARC-e & ARC-c & OBQA \\
\midrule
& Half-precision & $16 / 16$ & $76.33$ & $79.38$ & $75.35$ & $69.69$ & $65.78$ & $45.56$ & $44.00$ \\
\cmidrule{2-10}
LLaMA-$7$B & B+AdaRound & $4 / 16$ & $\mathbf{74.10}$ & $77.75$ & $73.60$ & $68.90$ & $57.79$ & $\mathbf{44.11}$ & $43.00$ \\
& B+FlexRound (Ours) & $4 / 16$ & $73.46$ & $\mathbf{78.35}$ & $\mathbf{74.43}$ & $\mathbf{69.14}$ & $\mathbf{63.43}$ & $43.43$ & $\mathbf{43.80}$ \\
\midrule
& Half-precision & $16 / 16$ & $81.90$ & $79.98$ & $78.41$ & $75.61$ & $70.79$ & $50.43$ & $47.20$ \\
\cmidrule{2-10}
LLaMA-$13$B & B+AdaRound & $4 / 16$ & $78.65$ & $79.54$ & $76.79$ & $\mathbf{75.53}$ & $63.38$ & $47.10$ & $45.20$ \\
& B+FlexRound (Ours) & $4 / 16$ & $\mathbf{78.78}$ & $\mathbf{79.71}$ & $\mathbf{77.40}$ & $75.30$ & $\mathbf{67.05}$ & $\mathbf{48.04}$ & $\mathbf{46.00}$ \\
\midrule
& Half-precision & $16 / 16$ & $85.96$ & $82.48$ & $82.20$ & $80.03$ & $74.87$ & $56.23$ & $47.00$ \\
\cmidrule{2-10}
LLaMA-$33$B & B+AdaRound  & $4 / 16$ & $84.65$ & $80.96$ & $80.03$ & $78.37$ & $67.51$ & $51.19$ & $44.60$ \\
& B+FlexRound (Ours) & $4 / 16$ & $\mathbf{86.64}$ & $\mathbf{81.83}$ & $\mathbf{81.26}$ & $\mathbf{79.01}$ & $\mathbf{70.66}$ & $\mathbf{53.24}$ & $\mathbf{45.00}$ \\
% \midrule
% & Half-precision & $85.96$ & $82.48$ & $82.20$ & $80.03$ & $74.87$ & $56.23$ & $47.00$ & \\
% \cmidrule{2-10}
% Five-shot  & Q+AdaRound & $85.96$ & $82.48$ & $82.20$ & $80.03$ & $74.87$ & $56.23$ & $47.00$ & \\
% & Q+FlexRound (Ours)& $\mathbf{85.32}$ & $\mathbf{80.90}$ & $\mathbf{80.52}$ & $\mathbf{78.37}$ & $\mathbf{71.72}$ & $\mathbf{53.16}$ & $\mathbf{46.80}$ & \\
\bottomrule
\end{tabular}
}
\end{center}
% \vskip -0.1in
\end{table*}

For all experiments, we employ the evaluation code from Eleuther AI's \textit{lm-evaluation-harness} \citep{eval-harness} for common sense reasoning bechmarks and the evaluation method in the Huggingface repository\footnote{\url{https://github.com/huggingface/transformers/tree/main/examples/pytorch/language-modeling}} for the causal language modeling task on WikiText2 without any modification. For reconstruction in all experiments, we use $512$ random samples from the training dataset of C4, and the number of iterations is set to $5000$. We use the Adam optimizer for all methods and models. For `Q + FlexRound' in Table~\ref{tab:plm_llama_zero_appendix} and Table~\ref{tab:plm_llama_five_appendix}, the batch size and the learning rate are set to $4$ and $3$e-$3$ for LLaMA-$7$B and LLaMA-$13$B, and $2$ and $1$e-$3$ for LLaMA-$33$B. For `B + FlexRound' in Table~\ref{tab:plm_llama_zero_appendix_w4a16} and Table~\ref{tab:plm_llama_five_appendix_w4a16}, the batch size and the learning rate are set to $4$ and $2$e-$4$ for LLaMA-$7$B, $4$ and $1$e-$4$ for LLaMA-$13$B, and $2$ and $1$e-$4$ for LLaMA-$33$B. The probability of dropping activation quantization is set to $0.5$ in the QDrop's setting. The other experimental setups of `B + AdaRound' and `Q + AdaRound' follow \citet{li2021brecq} and \citet{wei2022qdrop}, respectively.

\clearpage

\section{LLaMA fine-tuned via LoRA on WikiText2 and PTB}\label{appendix:llama_lora}

\begin{table*}[h]
% \vskip -5pt
% \vskip -0.1in
\caption{Performance of LLaMA-$7$B, LLaMA-$13$B, and LLaMA-$33$B fine-tuned via LoRA on WikiText2 and PTB, respectively. In LoRA, the query and value projection weights are adapted with a LoRA rank of $4$. The perplexity (PPL) is employed as a performance metric. The lower PPL, the better. ``Q $+$ X" means the implementation of X in the QDrop's setting. The weights of attention and feed-forward sub-layers are quantized to $8$-bit in a per-channel asymmetric format, whereas the input activations of those sub-layers are quantized to $8$-bit in a per-tensor asymmetric scheme.}\label{tab:llama_finetuned_lora}
\begin{center}
\small
% \resizebox{0.9\linewidth}{!}{
\begin{tabular}{clcccc}
\toprule
Dataset & \makecell{Method} & \# Bits (W/A) & LLaMA-$7$B & LLaMA-$13$B & LLaMA-$33$B \\
\midrule
& Half-precision (LoRA) & $16 / 16$ & $5.53$ & $5.07$ & $4.06$ \\
\cmidrule{2-6}
WikiText2 & Q+AdaRound & $8 / 8$ & $6.19$ & $5.80$ & $4.86$ \\
& Q+FlexRound (Ours) & $8 / 8$ & $\mathbf{5.73}$ & $\mathbf{5.29}$ & $\mathbf{4.32}$ \\
\midrule
& Half-precision (LoRA) & $16 / 16$ & $9.09$ & $8.47$ & $7.21$ \\
\cmidrule{2-6}
PTB & Q+AdaRound & $8 / 8$ & $9.85$ & $9.23$ & $8.21$ \\
& Q+FlexRound (Ours) & $8 / 8$ & $\mathbf{9.28}$ & $\mathbf{8.66}$ & $\mathbf{7.43}$ \\
\bottomrule
\end{tabular}
% }
\end{center}
% \vskip -10pt
% \vskip -0.15in
\end{table*}

\begin{table*}[h]
% \vskip -5pt
% \vskip -0.1in
\caption{Performance of LLaMA-$7$B, LLaMA-$13$B, and LLaMA-$33$B fine-tuned via LoRA on WikiText2 and PTB, respectively. In LoRA, the query and value projection weights are adapted with a LoRA rank of $4$. The perplexity (PPL) is employed as a performance metric. The lower PPL, the better. ``B $+$ X" implies the implementation of X in the BRECQ's setting. The weights of attention and feed-forward sub-layers are quantized to $3$-bit or $4$-bit in a per-channel asymmetric format, whereas the input activations of those sub-layers are kept in half-precision.}\label{tab:llama_finetuned_lora_w4a16}
\begin{center}
\small
% \resizebox{0.9\linewidth}{!}{
\begin{tabular}{clcccc}
\toprule
Dataset & \makecell{Method} & \# Bits (W/A) & LLaMA-$7$B & LLaMA-$13$B & LLaMA-$33$B \\
\midrule
& Half-precision (LoRA) & $16 / 16$ & $5.53$ & $5.07$ & $4.06$ \\
\cmidrule{2-6}
& B+AdaRound & $4 / 16$ & $5.72$ & $5.31$ & $4.33$ \\
WikiText2 & B+FlexRound (Ours) & $4 / 16$ & $\mathbf{5.63}$ & $\mathbf{5.14}$ & $\mathbf{4.17}$ \\
\cmidrule{2-6}
& B+AdaRound & $3 / 16$ & $6.41$ & $6.20$ & $4.98$ \\
& B+FlexRound (Ours) & $3 / 16$ & $\mathbf{5.88}$ & $\mathbf{5.33}$ & $\mathbf{4.40}$ \\

\midrule
& Half-precision (LoRA) & $16 / 16$ & $9.09$ & $8.47$ & $7.21$ \\
\cmidrule{2-6}
& B+AdaRound & $4 / 16$ & $9.27$ & $8.77$ & $7.35$ \\
PTB & B+FlexRound (Ours) & $4 / 16$ & $\mathbf{9.13}$ & $\mathbf{8.51}$ & $\mathbf{7.25}$ \\
\cmidrule{2-6}
& B+AdaRound & $3 / 16$ & $10.16$ & $8.98$ & $7.67$ \\
& B+FlexRound (Ours) & $3 / 16$ & $\mathbf{9.27}$ & $\mathbf{8.61}$ & $\mathbf{7.34}$ \\
\bottomrule
\end{tabular}
% }
\end{center}
% \vskip -10pt
% \vskip -0.15in
\end{table*}

For the LoRA configuration, we apply LoRA to the query and value projection weights with a LoRA rank of $4$. The batch size and the number of epochs are set to $128$ and $15$, respectively. For LLaMA-$7$B, LLaMA-$13$B, and LLaMA-$33$B, the learning rate is set to $1$e-$4$, $2$e-$4$, and $4$e-$5$ for Wikitext2 and $5$e-$4$, $4$e-$4$, and $6$e-$4$ for PTB.

For all experiments, we employ the evaluation method in the Huggingface repository\footnote{\url{https://github.com/huggingface/transformers/tree/main/examples/pytorch/language-modeling}} for WikiText2 and PTB without any modification. For reconstruction in all experiments, we use $256$ random samples from the training dataset of WikiText2 and PTB respectively, and the number of iterations is set to $5000$. We use the Adam optimizer for all methods and models. For the experiments of `Q + FlexRound' on WikiText2 in Table~\ref{tab:llama_finetuned_lora}, the batch size and the learning rate are set to $4$ and $5$e-$3$ for LLaMA-$7$B, $4$ and $2$e-$3$ for LLaMA-$13$B, and $2$ and $2$e-$3$ for LLaMA-$33$B. For the experiments of `Q + FlexRound' on PTB in Table~\ref{tab:llama_finetuned_lora}, the batch size and the learning rate are set to $4$ and $2$e-$3$ for LLaMA-$7$B, $4$ and $1$e-$3$ for LLaMA-$13$B, and $2$ and $3$e-$3$ for LLaMA-$33$B. 
For the experiments of `B + FlexRound' with $4$-bit weight quantization on WikiText2 in Table~\ref{tab:llama_finetuned_lora_w4a16}, the batch size and the learning rate are set to $4$ and $5$e-$4$ for LLaMA-$7$B and LLaMA-$13$B, and $2$ and $2$e-$4$ for LLaMA-$33$B. For the experiments of `B + FlexRound' with $4$-bit weight quantization on PTB in Table~\ref{tab:llama_finetuned_lora_w4a16}, the batch size and the learning rate are set to $4$ and $5$e-$4$ for LLaMA-$7$B and LLaMA-$13$B, and $2$ and $1$e-$3$ for LLaMA-$33$B. 
For the experiments of `B + FlexRound' with $3$-bit weight quantization on WikiText2 in Table~\ref{tab:llama_finetuned_lora_w4a16}, the batch size and the learning rate are set to $4$ and $3$e-$4$ for LLaMA-$7$B and LLaMA-$13$B, and $2$ and $3$e-$4$ for LLaMA-$33$B. For the experiments of `B + FlexRound' with $3$-bit weight quantization on PTB in Table~\ref{tab:llama_finetuned_lora_w4a16}, the batch size and the learning rate are set to $4$ and $7$e-$4$ for LLaMA-$7$B, $4$ and $6$e-$4$ for LLaMA-$13$B, and $2$ and $6$e-$4$ for LLaMA-$33$B.
The probability of dropping activation quantization is set to $0.5$ in the QDrop's setting. The other experimental setups of `B + AdaRound' and `Q + AdaRound' follow \citet{li2021brecq} and \citet{wei2022qdrop}, respectively.

\end{document}